%% file: iclr2027_conference.tex
\documentclass{article}
\usepackage{xcolor}
\usepackage{colortbl}
\usepackage{tikz}
\usetikzlibrary{calc}
\usepackage{soul}
\usepackage{iclr2027_conference,times}

\usepackage{hyperref}
\usepackage{url}
\usepackage{pifont}

\usepackage{amsmath,amssymb}
\usepackage{algorithm}
\usepackage{algpseudocode}
\usepackage{booktabs}
\usepackage{array,tabularx}
\usepackage{xspace}
\usepackage{listings}

\definecolor{baseTaskColor}{HTML}{245A4A}      %
\definecolor{novelTaskColor}{HTML}{D9A928}     %
\definecolor{disclosedTaskColor}{HTML}{9A6A45} %

\definecolor{agniCharacterizeColor}{HTML}{2F6B5F} %
\definecolor{agniSynthesizeColor}{HTML}{D4A72C}   %
\definecolor{agniInstantiateColor}{HTML}{58758A}  %
\definecolor{agniValidateColor}{HTML}{B45F3C}     %
\definecolor{promptHeader}{RGB}{220,220,220}

\newcommand{\agnistage}[3]{%
    \Statex
    \ifnum#3=1
        \tikz[remember picture,baseline] \coordinate (#1-start) at (0,-0.10em);%
        \tikz[remember picture,overlay] \coordinate (agni-stage-left) at (0,0);%
    \else
        \tikz[remember picture,baseline] \coordinate (#1-start);%
    \fi
    \hfill
    \ifnum#3=1
        \tikz[remember picture,overlay] \coordinate (agni-stage-right) at (0,0);%
    \fi
    \ifnum#3=1
        \vspace{-0.65\baselineskip}%
    \else
        \vspace{-\baselineskip}%
    \fi
}
\newcommand{\agniendstage}[2]{%
    \tikz[remember picture,overlay] \coordinate (#1-end) at (0,-0.45em);%
    \tikz[remember picture,overlay]{%
        \fill[#2,fill opacity=0.09]
        ($(agni-stage-left|-#1-start)+(-0.35em,1.00em)$) rectangle
        ($(agni-stage-right|-#1-end)+(0.35em,0.20em)$);%
        \shade[
            shading=axis,
            top color=#2,
            middle color=white,
            bottom color=#2,
            shading angle=90,
            opacity=0.05
        ]
        ($(agni-stage-left|-#1-start)+(-0.35em,1.00em)$) rectangle
        ($(agni-stage-right|-#1-end)+(0.35em,0.20em)$);%
    }%
}

\DeclareRobustCommand{\agnicaptionstage}[2]{%
    {\sethlcolor{#1!10}\hl{#2}}%
}

\usepackage[most]{tcolorbox}
\usepackage{multirow}
\usepackage{wrapfig}
\AddToHook{env/wrapfigure/before}{\setlength{\intextsep}{3pt}}
\AddToHook{env/wraptable/before}{\setlength{\intextsep}{3pt}}
\makeatletter
\newcommand{\finishcompactwrap}{%
    \par
    \ifnum\c@WF@wrappedlines>1
        \begingroup
        \@tempcnta=\c@WF@wrappedlines
        \advance\@tempcnta by -1
        \dimen@=\baselineskip
        \multiply\dimen@ by \@tempcnta
        \vskip\dimen@
        \endgroup
    \fi
    \WFclear
}
\newcommand{\fitcompactwrap}{%
    \par\penalty0 %
    \ifdim\WF@size>\dimexpr\pagegoal-\pagetotal\relax
        \newpage
    \fi
}
\makeatother

\usepackage{placeins}
\usepackage{float}
\usepackage{titletoc}

\usepackage{cleveref}

\newcommand{\promptheading}[1]{%
    \textbf{#1}\par%
}
\newcommand{\promptbullet}{%
    \raisebox{0.08ex}{\normalsize\bfseries\textbullet}\hspace{0.25em}%
}
\newtcblisting[auto counter, number within=section]{promptbox}[2][]{
    enhanced jigsaw,
    breakable,
    listing only,
    colback=white,
    colbacktitle=promptHeader,
    colframe=black!70,
    coltitle=black,
    boxrule=0.5pt,
    arc=0pt,
    outer arc=0pt,
    left=8pt,
    right=8pt,
    top=7pt,
    bottom=7pt,
    lefttitle=8pt,
    righttitle=8pt,
    toptitle=4pt,
    bottomtitle=4pt,
    boxsep=0pt,
    before skip=18pt,
    after skip=14pt,
    fonttitle=\sffamily\small\bfseries,
    title={Prompt~\thetcbcounter: #2},
    title after break={Prompt~\thetcbcounter: #2 (continued)},
    extras middle={colbacktitle=promptHeader!25!white,coltitle=black!40},
    extras last={colbacktitle=promptHeader!25!white,coltitle=black!40},
    label={#1},
    listing options={
        basicstyle=\ttfamily\small,
        breaklines=true,
        breakatwhitespace=true,
        breakindent=0pt,
        breakautoindent=false,
        columns=fullflexible,
        keepspaces=true,
        showstringspaces=false,
        escapeinside={(*@}{@*)},
        literate={* }{{\promptbullet}}2,
        emph={CORE,IDEA,SUBTLE,NOTICE,ADAPT,COVERT,NEVER,ONLY,MUST,JSON,ROBUSTNESS,DEFAULT,ASSUMPTION,DISCOVER,BITE,MOVE,FIND,ALREADY,SILENTLY,DIFFERENT},
        emphstyle=\bfseries,
        aboveskip=0pt,
        belowskip=0pt
    }
}
\makeatletter
\renewcommand{\thetcb@cnt@promptbox}{P.\arabic{tcb@cnt@promptbox}}
\makeatother

\title{When Successful Strategies Fail: Adaptation to Environmental Novelty in Terminal Agents}

\newcommand{\authormarkheight}{5pt}
\newcommand{\uiuclogo}[1][0.8em]{\includegraphics[height=#1]{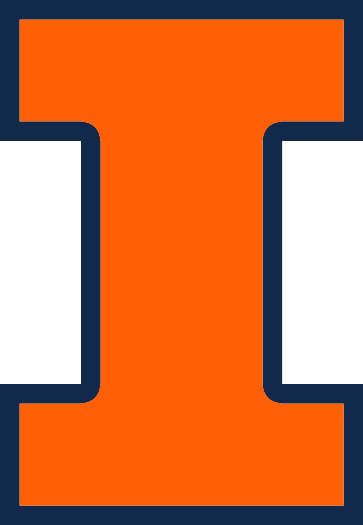}}
\newcommand{\mslogo}[1][0.8em]{\includegraphics[trim=0.72bp 0.72bp 0.72bp 0.72bp,clip,height=#1]{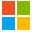}}
\author{\textbf{Janvijay Singh\kern0.1em\textsuperscript{\uiuclogo[\authormarkheight]\,\mslogo[\authormarkheight]}\thanks{Work done as a Microsoft Research intern. Correspondence: \href{mailto:jvsingh2@illinois.edu}{jvsingh2@illinois.edu}.}\hspace{0.7em}, Vaishnavi Shrivastava\kern0.1em\textsuperscript{\mslogo[\authormarkheight]}} \\
\textbf{Dilek Hakkani-T\"ur\kern0.1em\textsuperscript{\uiuclogo[\authormarkheight]}, Ece Kamar\kern0.1em\textsuperscript{\mslogo[\authormarkheight]}, Asli Celikyilmaz\kern0.1em\textsuperscript{\mslogo[\authormarkheight]}} \\
\uiuclogo\ University of Illinois Urbana-Champaign \\
\mslogo\ Microsoft Research AI Frontiers
}

\begin{document}
\raggedbottom

\begingroup
\makeatletter
\renewcommand{\@fnsymbol}[1]{\hbox{\raisebox{-1.55pt}{\resizebox{!}{6.5pt}{\ding{41}}}}}
\makeatother
\maketitle
\endgroup

\begin{abstract}

LLM agents increasingly solve long-horizon tasks by autonomously interacting with their environment. In doing so, their strategies rely on assumptions about that environment: which resources and tools exist, where they are located, and how they behave. When these assumptions no longer hold, reliable agents must detect the change and adapt while pursuing the same goal. We study this \textit{adaptation capability} through \textit{environmental novelty}: a change that keeps the task objective fixed while invalidating an assumption underlying an otherwise successful trajectory. We introduce AGNI, an automated pipeline that extracts trajectory-relevant assumptions, injects targeted environmental changes, and validates that the resulting novel tasks remain solvable. Across three terminal benchmarks, AGNI produces diverse novelties spanning resources, interfaces, constraints, and execution semantics. Evaluating multiple LLM agents reveals a substantial adaptation gap between base and novel tasks. Trajectory analysis suggests that agents often encounter evidence of the change but fail to diagnose its cause and revise their strategy. Finally, post-training for environmental novelty improves adaptation to held-out novel tasks while also improving performance on base tasks. Our results highlight a gap between task competence and adaptive capability and motivate environmental variation as a core dimension of agent training and evaluation.

\end{abstract}

\input{sections/introduction}

\input{sections/related_works}

\input{sections/env_novelty}

\input{sections/agni-temp}

\input{sections/evaluating_adaptation}

\input{sections/post_training_adaptation}

\input{sections/discussion}

\input{sections/conclusion}

\input{references}
\clearpage
\appendix

\startcontents[appendix]
\section*{Appendix Contents}

\printcontents[appendix]{}{1}{}
\clearpage

\input{sections/appendix/statements}

\input{sections/appendix/detailed-related-work}

\input{sections/appendix/agni-details}

\input{sections/appendix/agni-details/b4-realism-analysis}

\input{sections/appendix/detailed-evaluation-results}

\input{sections/appendix/agni-details/b2-trajectory-annotations}

\input{sections/appendix/post-training-details}

\input{sections/appendix/agni-details/c-prompts-and-output-schemas}

\end{document}

%% file: sections/introduction.tex
\section{Introduction}

LLM agents increasingly pursue long-horizon tasks from high-level goals rather than procedural instructions~\citep{Yang2024SWEagentAI,Xie2024OSWorldBM}.
This increased autonomy requires agents to construct and execute task-solving strategies.
These strategies often rely on agents' implicit assumptions about their environment, such as available resources, interface behavior, and other agents' behavior.
Since these assumptions are often unspecified by the task, whether they hold may only become clear through interaction.
Reliable agents must, therefore, go beyond executing an initial strategy and adapt when their assumptions fail.
They must identify the relevant environmental changes, revise their strategy, and continue pursuing the original goal~\citep{Langley2020OpenWorldLF,Kejriwal2024ChallengesEA}.

We study this adaptation capability through environmental novelty: a change in the environment that invalidates an assumption used by a successful trajectory, while preserving the task objective.
We particularly focus on terminal agents for two reasons.
First, terminal agents span heterogeneous tasks, from simple file processing to advanced machine learning, and can interact with open-ended tools and resources~\citep{merrill2026terminalbench}.
This diversity makes them a suitable testbed for studying whether LLMs can reason effectively to adapt across diverse strategies and environments.
Second, adaptation in software environments is practically important, since
software continually evolves through changes in interfaces, dependencies, configurations, and workflows~\citep{dig2006apis}.

We formalize environmental novelty by separating the task objective from the software environment in which it is achieved.
Given a base terminal task and a verified successful trajectory, the trajectory may rely on assumptions not specified by the task instruction.
These assumptions can include a writable directory, an installed dependency version, or a particular tool interface.
An environmental novelty modifies the environment so that such an assumption no longer holds, while preserving the instruction, success criterion, and solvability.
Since the novelty is undisclosed to the agent, it must infer the changed condition through interaction, revise its strategy, and complete the original task.
The performance gap between base and novel tasks reflects a model's ability to adapt.

Terminal tasks are well suited for studying adaptation, but their heterogeneity and open-endedness also make base--novel pairs difficult to construct.
A task can often be solved using a diverse set of command sequences and resources.
As a result, the relevant environmental assumptions are often implicit and cannot be inferred from the task instructions alone.
Arbitrary changes in the environment may leave a successful strategy unaffected, admit trivial workarounds, or break task solvability.
We therefore introduce \textsc{AGNI} (Assumption-Guided Novelty Injection), an LLM-based pipeline that starts from verified successful trajectories, identifies the environmental assumptions supporting their strategies, and invalidates one while preserving the task objective and solvability.
\textsc{AGNI} retains variants where the original strategy no longer suffices but an adapted trajectory can succeed, enabling controlled evaluation, trajectory-level analysis, and post-training under environmental novelty.

We study adaptation by evaluating seven diverse LLMs on \textsc{AGNI}-generated task pairs.
We generate task pairs from three terminal datasets of varying difficulty: Endless Terminals, TB-Lite, and TB-2.
We find that models consistently degrade on novel tasks, and higher base-task performance does not reliably predict adaptation capability.
We further test whether increased test-time reasoning improves adaptation and find that it often helps, but not consistently across models.
We also analyze whether disclosing the environmental change in the task instruction recovers the adaptation gap, and find that it recovers a substantial fraction across models.
However, this recovery cannot be attributed simply to whether agents observe the novelty during interaction.
Our trajectory-level analysis shows that stronger adapters are better at turning observed novelty into a diagnosis and a revised strategy, while weaker adapters more often repeat failed approaches and terminate early.

We next ask whether post-training can improve adaptation to novelty.
Standard agentic post-training~\citep{Shao2024DeepSeekMathPT,Shrivastava2026ECHOTA}
improves performance on both base and novel terminal tasks, but leaves a significant adaptation gap.
We develop two complementary approaches to narrow this gap.
First, we augment training with novel environments that require agents to solve familiar tasks under changed conditions~\citep{Devries2017ImprovedRO,Park2019SpecAugmentAS,Liu2026PayingLG}.
Second, we introduce explicit \textit{meta-action} reasoning, where agents reason about the strategic intent of each action, such as inspecting, diagnosing, revising, or verifying.
Both novelty-augmented training and meta-action reasoning provide improvements on held-out novel tasks, with gains transferring to an additional terminal benchmark.
We also study process rewards over meta-actions, finding that they reshape agent behavior but provide limited improvement in adaptation.

Overall, our work highlights adaptation as a key capability for reliable autonomous agents.
We make three contributions.
First, we formalize environmental novelty as the controlled invalidation of strategy-relevant assumptions and introduce \textsc{AGNI}, a scalable pipeline for injecting such novelties into terminal tasks.
Second, we use \textsc{AGNI}-generated environments to study adaptation and how it unfolds during agentic interaction.
Third, we develop post-training approach based on data augmentation and meta-action reasoning to improve adaptation.
Together, our results motivate environmental variation as a core axis for training and evaluating terminal agents.

%% file: sections/related_works.tex
\section{Related Work}

\textbf{Open-World Learning and Adaptation.}
Open-world learning studies how agents adapt to environmental novelty~\citep{Langley2020OpenWorldLF,Kejriwal2024ChallengesEA}.
Prior work decomposes adaptation into novelty detection, diagnosis, and behavioral updates~\citep{Muhammad2021ANA,Loyall2025ColtraneAD}.
Crucially, novelty is relative to the agent and task: an environmental change may invalidate one strategy while being irrelevant to another~\citep{Boult2021TowardsAU,Goel2024ANC}.
Prior work largely focuses on embodied, simulated, or game environments with structured world models and adaptation components.
In contrast, we study general-purpose LLM agents operating through open-ended programmatic interfaces, where strategies and assumptions are implicit in interaction trajectories.

\textbf{Robustness Evaluation in LLM Agents.}
Recent work evaluates LLM-agent robustness under perturbations to tool
availability~\citep{Liu2026PlanBenchXLEL,liu-etal-2026-costbench}, tool
interfaces~\citep{Tian2026BeyondFC,wu2026can}, workspace
contents~\citep{Li2026RepoMiragePR,Mahmud2026AJF}, and runtime
conditions~\citep{Jha2026AgentMT}.
These works largely predefine perturbation families.
\textsc{AGNI} instead derives perturbations from assumptions revealed by
successful trajectories, making novelty strategy-dependent.
This is important for terminal agents, where heterogeneous tasks and open-ended
tools and resources make relevant environmental dependencies difficult to
enumerate in advance.
Furthermore, since \textsc{AGNI} targets a known assumption, it enables
a finer-grained analysis of adaptation within the interaction trajectory.

\textbf{Post-training for Adaptive LLM Agents.}
Recent work post-trains LLM agents to recover from perturbations in tool-use~\citep{Vuddanti2025PALADINSL,wu2026can}, text-game~\citep{Ye2026LookBY}, and GUI environments~\citep{sun2026agenthijack}.
These works typically train on predefined perturbations and supervise recovery using task outcomes, recovery demonstrations, supplied diagnoses, and incentives for reflection and exploration.
In contrast, we train on trajectory-grounded perturbations and provide process-level adaptation guidance through self-reported and rewarded meta-actions, including inspecting, diagnosing, revising, and verifying.

\textbf{Software Evolution and Resilience Testing.}
Software routinely evolves in ways that break previously valid workflows. Interfaces and dependencies change~\citep{dig2006apis,kula2018dependencies,pep668}, configuration errors arise~\citep{yin2011configuration,xu2015configuration}, and system state changes can also expose failures~\citep{yuan2014simpletesting}. These observations motivate our study of adaptation to environmental novelty. Analogous to resilience testing in software systems~\citep{basiri2016chaos}, we study whether terminal agents remain reliable as their software environment changes.

A more detailed discussion of related work and additional references is provided in Appendix~\ref{app:detailed-related-work}.

%% file: sections/env_novelty.tex
\begin{figure}[t]
    \centering
    \includegraphics[width=\linewidth]{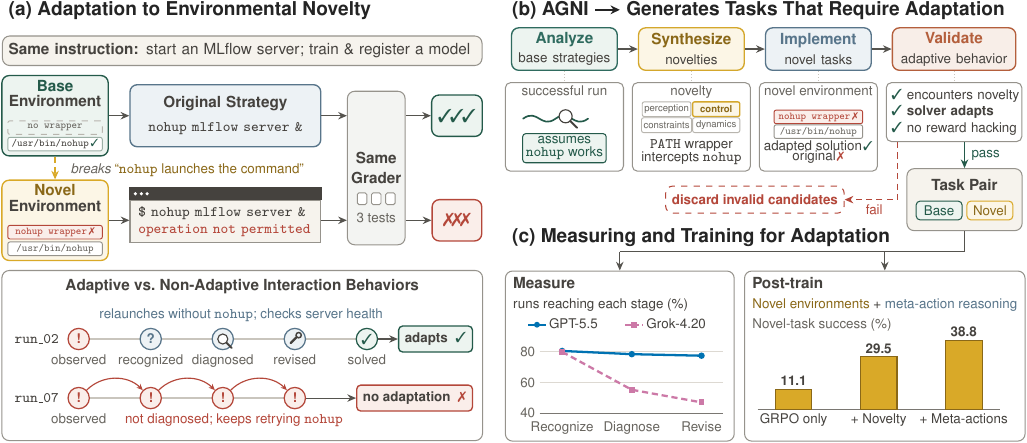}
    \caption{
    \textbf{Adaptation to Environmental Novelty in Terminal Agents.}
    \textbf{(a)} Novelty can disrupt an agent's original strategy and recovery depends on adaptive behaviors during interaction (example from Appendix~\ref{app:agni-examples}).
    \textbf{(b)} \textsc{AGNI} injects novelties into base environments such that task success requires adaptation, while preserving task objective and solvability.
    \textbf{(c)} We use \textsc{AGNI}-generated tasks to measure adaptation, analyze adaptive behaviors, and post-train agents to improve adaptation.
    }
    \label{fig:overview}
\end{figure}

\section{Environmental Novelty and Adaptation in Terminal Tasks}
\label{sec:environmental-novelty}

We now formalize environmental novelty and adaptation in terminal tasks.

\textbf{Terminal Task.}
We represent a terminal task as
\(\mathcal{T}=(I,E,G)\),
where \(I\) is the instruction, \(E\) is the execution environment, typically a Docker container, and \(G\) is a task-specific outcome grader implemented with deterministic evaluation scripts.
Given \(\mathcal{T}\), an agent executes a task-solving strategy through interaction with \(E\), producing a trajectory
$\tau=(o_0,r_1,a_1,o_1,\ldots,r_T,a_T,o_T)$,
where \(o_0\) contains \(I\) and the initial observation, and \(r_t\), \(a_t\), and \(o_t\) denote reasoning, action, and resulting observation at turn \(t\).
The task succeeds if \(G(\tau,E)=1\) and fails otherwise.
This formulation separates what the agent must accomplish, specified by \(I\) and \(G\), from the environmental conditions under which it must do so, specified by \(E\).

\textbf{Environmental Novelty.}
A successful trajectory \(\tau\) may depend on environmental assumptions that are not specified by \(I\), reflecting the agent’s learned priors about how such environments typically behave.
Let \(\mathcal{A}(\tau,E)\) denote the set of trajectory-relative environmental assumptions on which the success of \(\tau\) depends, and let \(\alpha \in \mathcal{A}(\tau,E)\) be one such assumption.
An \emph{environmental novelty} transforms \(E\) into \(E'\) such that \(\alpha\) no longer holds, while preserving the task objective and success criterion.
For example, \(\tau\) may write an artifact to \texttt{/workspace} while implicitly assuming that the directory is writable. A novelty may make this directory read-only.
Thus, the strategy underlying \(\tau\) succeeds in \(E\) but is no longer sufficient in \(E'\).

\textbf{Adaptation.}
Once an environmental assumption is invalidated, an adaptable agent must revise its strategy while still achieving the task objective. This may require recognizing evidence of novelty, diagnosing which assumption has been invalidated, and revising the strategy appropriately. We focus on the autonomous setting, where the agent must discover the environmental change through interaction and recover without explicit human guidance about the novelty or how to respond to it.
Accordingly, we define adaptation as \emph{task-consistent recovery}: revising
the affected strategy in response to the changed environmental condition and
successfully completing the original task, without requiring any particular
recovery strategy.
In the preceding example, the agent could adapt by restoring write access
to \texttt{/workspace} and then completing the original task.

To measure adaptation, we compare an agent's performance across three task variants:
\[
\mathcal{T}_{\mathrm{base}}=(I,E,G), \qquad
\mathcal{T}_{\mathrm{novel}}=(I,E',G), \qquad
\mathcal{T}_{\mathrm{disc}}=(I_{\mathrm{disc}},E',G).
\]
A \emph{controlled task pair} $(\mathcal{T}_{\mathrm{base}},\mathcal{T}_{\mathrm{novel}})$
consists of the original environment \(E\) and a solvable modified environment
\(E'\) that invalidates an assumption underlying a verified successful trajectory in \(E\).
The pair preserves the instruction and success criterion while changing an
assumption relied upon by \(\tau\).
In the disclosed variant, \(I_{\mathrm{disc}}\) explicitly states the novelty but provides no guidance on how to adapt.

Let \(P_{\mathrm{base}}\), \(P_{\mathrm{novel}}\), and \(P_{\mathrm{disc}}\) denote the agent's pass@1 rates on the three variants. We define the \emph{adaptation gap} \(\Delta_{\mathrm{adapt}}\), \emph{disclosure gain} \(\Delta_{\mathrm{disc}}\), and \emph{recovery fraction} \(\rho_{\mathrm{rec}}\) as
\[
\Delta_{\mathrm{adapt}} = P_{\mathrm{base}} - P_{\mathrm{novel}},
\qquad
\Delta_{\mathrm{disc}} = P_{\mathrm{disc}} - P_{\mathrm{novel}},
\qquad
\rho_{\mathrm{rec}} = \Delta_{\mathrm{disc}} / \Delta_{\mathrm{adapt}}.
\]
A larger adaptation gap indicates greater degradation under environmental novelty, while a smaller gap suggests better adaptation.
We interpret this gap alongside base and novel success rates to assess a model's adaptive capability, since the gap also depends on base-task competence.
By contrast, disclosure gain measures the performance recovered when the novelty is made explicit.
This gain helps distinguish the difficulty of identifying the relevant environmental change through interaction from the difficulty of adapting once it is known.
For a positive adaptation gap, \(\rho_{\mathrm{rec}}\) measures the fraction of lost performance recovered through disclosure.

%% file: sections/agni-temp.tex
\input{sections/agni-algo}

\section{AGNI: Assumption-Guided Novelty Injection}
\label{sec:agni}

Terminal tasks are heterogeneous across domains and open-ended in the tools, resources, and strategies available to agents.
This diversity makes enumerating novelties from the task instruction alone difficult; and an arbitrary environmental change may be irrelevant to the strategy or make the task unsolvable.
We therefore introduce \textsc{AGNI}, a scalable LLM-based pipeline that injects strategy-relevant novelties while preserving task solvability.
Figure~\ref{fig:overview} illustrates \textsc{AGNI}, Algorithm~\ref{alg:agni} presents the procedure, and Appendix~\ref{app:agni-details} provides full details.
We briefly describe the key components below.

\textbf{Generating Trajectory-Relative Novelties.}
For each base task, \textsc{AGNI} samples multiple trajectories from a capable \textit{solver} model and retains tasks with at least one successful trajectory.
Then, an \textit{analyzer} model examines the base task $(I,E,G)$ and sampled trajectories to characterize solver strategies and underlying environmental assumptions.
Using these assumptions, \textsc{AGNI} identifies applicable novelty dimensions and generates candidate novelty descriptions that target the corresponding environmental conditions.
We consider four novelty dimensions: what the agent observes (\emph{perception}), how it can act (\emph{control}), what actions are permitted (\emph{constraints}), and how the environment responds (\emph{dynamics}).
\textsc{AGNI} then de-duplicates and filters candidates for viability, as assessed by the analyzer.
It then implements and validates the corresponding environments, as described next.

\textbf{Implementing and Validating Novel Tasks.}
Building on prior work on terminal task generation~\cite{Gandhi2026EndlessTS,Ivison2026TmaxAS}, \textsc{AGNI} implements each novelty by modifying the base environment $E$ to obtain $E'$ while keeping the instruction $I$ and grader $G$ unchanged.
It also generates an adapted oracle solution for the novel variant.
\textsc{AGNI} retains a novelty implementation only when the adapted oracle solution succeeds and the original oracle solution fails.
This ensures that the task remains solvable while the novelty invalidates the base strategy.
\textsc{AGNI} then samples solver trajectories in $E'$ and retains tasks with at least one successful trajectory.
Finally, it analyzes successful trajectories turn by turn to verify that the solver encounters the novelty and adapts to it.
We also reject cases where the solver exhibits reward hacking.
Together, these checks verify solvability and provide evidence of legitimate adaptation in sampled successful trajectories.

\input{sections/tables/agni-question-1}

\textbf{Curating the Final Datasets.}
We run \textsc{AGNI} on three base-task sources: Endless Terminals (ET)~\citep{Gandhi2026EndlessTS}, TB-Lite~\citep{OpenThoughts-TBLite}, and Terminal-Bench 2 (TB-2)~\citep{merrill2026terminalbench}.
We generate base--novel task pairs for evaluation from all three sources and use ET to construct the training set at scale, as TB-Lite and TB-2 are evaluation-only datasets.
For evaluation, we construct evaluation sets through joint stratification across dimensions capturing difficulty and real-world plausibility.
We stratify by base-task performance $P_{\mathrm{base}}$, relative novelty-induced difficulty
$\delta_{\mathrm{adapt}}=(P_{\mathrm{base}}-P_{\mathrm{novel}})/P_{\mathrm{base}}$,
and real-world plausibility.
We estimate $P_{\mathrm{base}}$ and $P_{\mathrm{novel}}$ using a weak solver model and use an LLM judge to score whether each novelty could plausibly occur in real-world deployments.
This yields 4,000 training pairs from ET and evaluation sets of 140 ET-eval, 118 TB-Lite, and 87 TB-2 pairs.

\textbf{Implementation Details.}
All \textsc{AGNI} model components use GPT-5.5, with high reasoning effort for the solver and medium effort otherwise.
We allocate higher effort to the solver because its capability bounds the difficulty of retained tasks.
More details in Appendix~\ref{app:agni-details}.

%% file: sections/agni-algo.tex
\begin{figure}[!t]
\centering
\scalebox{1.0}{%
\begin{minipage}{\linewidth}
\setlength{\intextsep}{0pt}
\begin{algorithm}[H]
\caption[\textsc{AGNI}: Assumption-Guided Novelty Injection]{%
\textsc{\textbf{AGNI}}: \textbf{A}ssumption-\textbf{G}uided \textbf{N}ovelty \textbf{I}njection.
\textsc{AGNI} generates validated base--novel task pairs in four stages by:
\agnicaptionstage{agniCharacterizeColor}{ (a) analyzing base-task strategies};
\agnicaptionstage{agniSynthesizeColor}{ (b) synthesizing novelty descriptions};
\agnicaptionstage{agniInstantiateColor}{ (c) implementing and validating novel tasks}; and
\agnicaptionstage{agniValidateColor}{(d) validating adaptive behavior}.}
\label{alg:agni}
\centering
\begin{minipage}{\linewidth}
\begin{algorithmic}[1]

\Require Base task \(\mathcal{T}_{\mathrm{base}}=(I,E,G)\);
oracle solution \(S\); solver model \(M_{\mathrm S}\);
analyzer model \(M_{\mathrm A}\); novelty injector model \(M_{\mathrm N}\);
discarded-novelty history \(\mathcal{H}\);
trajectory sampling budget \(K\); novelty specification budget \(N_{\mathrm{spec}}\);
retention budget \(N_{\mathrm{keep}}\)

\Ensure Validated task pairs
\(\mathcal{V}=\{(\mathcal{T}_{\mathrm{base}},\mathcal{T}_{\mathrm{novel}})\}\)

\agnistage{agni-characterize}{Characterize Base Strategy}{1}

\State
\(\mathcal{R}_{\mathrm{base}}
\gets
\{(\tau_{\mathrm{base}}^{(k)},G(\tau_{\mathrm{base}}^{(k)},E))
:\tau_{\mathrm{base}}^{(k)}\sim M_{\mathrm S}(I,E)\}_{k=1}^{K}\)
\Comment{sample and grade base trajectories}

\State
\textbf{if}
\(\nexists\,(\tau,r)\in\mathcal{R}_{\mathrm{base}}\)
such that \(r=1\),
\textbf{return} \(\emptyset\)
\Comment{ensure base-task solvability}

\State
\((\mathcal{B},\mathcal{A})
\gets
M_{\mathrm A}\!\left(
\mathcal{T}_{\mathrm{base}},S,\mathcal{R}_{\mathrm{base}}
\right)\)
\Comment{extract resource-use behavior and assumptions}

\agniendstage{agni-characterize}{agniCharacterizeColor}
\agnistage{agni-synthesize}{Synthesize Novelty Candidates}{0}

\State
\(\mathcal{D}
\gets
M_{\mathrm N}^{\mathrm{dim}}
(\mathcal{T}_{\mathrm{base}},\mathcal{B},\mathcal{A})\)
\Comment{identify applicable novelty dimensions}

\For{\(d \in \mathcal{D}\)}
    \State
    \(\mathcal{C}_d
    \gets
    M_{\mathrm N}^{\mathrm{spec}}
    (\mathcal{T}_{\mathrm{base}},\mathcal{B},\mathcal{A},
     d,\mathcal{H}_d,N_{\mathrm{spec}})\)
    \Comment{generate novelty descriptions}
\EndFor

\State
\(\mathcal{C}
\gets
\operatorname{Select}\!\left(
    \operatorname{Filter}\!\left(
        \textstyle\bigcup_{d\in\mathcal{D}}\mathcal{C}_d
    \right)
\right)\)
\Comment{de-duplicate and filter by viability}

\agniendstage{agni-synthesize}{agniSynthesizeColor}
\agnistage{agni-instantiate}{Instantiate \& Validate Novel Task}{0}

\State
\(\mathcal{V}\gets\emptyset\)

\For{\(c \in \mathcal{C}\)}

    \State
    \((E'_c,S'_c)
    \gets
    M_{\mathrm N}^{\mathrm{impl}}
    (\mathcal{T}_{\mathrm{base}},S,c)\)
    \Comment{implement novelty and adapted solution}

    \State
    \textbf{if}
    \(G(S'_c,E'_c)=0\),
    \textbf{continue}
    \Comment{ensure novel-task solvability}

    \State
    \textbf{if}
    \(G(S,E'_c)=1\),
    \textbf{continue}
    \Comment{ensure novelty invalidates the base strategy}

    \State
    \(\mathcal{R}_{c}
    \gets
    \{(\tau_{c}^{(k)},G(\tau_{c}^{(k)},E'_c))
    :\tau_{c}^{(k)}\sim M_{\mathrm S}(I,E'_c)\}_{k=1}^{K}\)
    \Comment{sample and grade novel trajectories}

    \State
    \textbf{if}
    \(\nexists\,(\tau,r)\in\mathcal{R}_{c}\)
    such that \(r=1\),
    \textbf{continue}
    \Comment{ensure solver can succeed on novel task}

    \agniendstage{agni-instantiate}{agniInstantiateColor}
    \agnistage{agni-adaptive}{Validate Adaptive Behavior}{0}

    \State
    \(v_c
    \gets
    M_{\mathrm A}^{\mathrm{obs}}
    (c,\mathcal{R}_{\mathrm{base}},\mathcal{R}_{c})\)
    \Comment{evaluate expected trajectory change}

    \State
    \textbf{if} \(v_c=0\),
    \textbf{continue}
    \Comment{verify the expected trajectory-level change}

    \State
    \(\mathcal{L}_{c}
    \gets
    \Call{AnalyzeTrajectories}{c,\mathcal{R}_{c}}\)
    \Comment{label novelty handling stages per trajectory}

    \State
    \textbf{if}
    \(\neg\forall\,\ell\in\mathcal{L}_{c}:
    \operatorname{Solved}(\ell)\Rightarrow\operatorname{Adapt}(\ell)\),
    \textbf{continue}
    \Comment{ensure successful trajectories adapt}

    \State
    \textbf{if}
    \(\exists\,\ell\in\mathcal{L}_{c}:
    \operatorname{RewardHack}(\ell)\),
    \textbf{continue}
    \Comment{reject observed reward hacking}

    \State
    \(\mathcal{T}'_c\gets(I,E'_c,G)\)

    \State
    \(\mathcal{V}
    \gets
    \mathcal{V}\cup
    \{(\mathcal{T}_{\mathrm{base}},\mathcal{T}'_c)\}\)
    \Comment{retain validated task pair}

    \State
    \textbf{if}
    \(|\mathcal{V}|=N_{\mathrm{keep}}\),
    \textbf{break}

\EndFor

\State \Return \(\mathcal{V}\)
\agniendstage{agni-adaptive}{agniValidateColor}

\end{algorithmic}
\end{minipage}
\end{algorithm}
\end{minipage}%
}
\end{figure}

%% file: sections/tables/agni-question-1.tex
\newcommand{\AgniTableOneWidth}{1.0\textwidth}

\begin{table}[!t]
\centering
\resizebox{\AgniTableOneWidth}{!}{%
\begin{tabular}{l|ccc|ccc|ccc}
\toprule
& \multicolumn{3}{c|}{ET-eval}
& \multicolumn{3}{c|}{TB-Lite}
& \multicolumn{3}{c}{TB-2} \\
Model
& Base & Novel & $\Delta_{\mathrm{adapt}}$
& Base & Novel & $\Delta_{\mathrm{adapt}}$
& Base & Novel & $\Delta_{\mathrm{adapt}}$ \\
\midrule
GPT-5.5
& 97.0 & 74.5 & 22.5
& 93.2 & 71.7 & 21.5
& 88.8 & 79.7 & 9.1 \\

GPT-5.4-mini
& 94.5 & 61.9 & 32.6
& 65.0 & 39.5 & 25.5
& 75.9 & 46.6 & 29.3 \\

Kimi-2.6
& 92.7 & 74.7 & 18.0
& 83.6 & 58.3 & 25.3
& 73.8 & 61.4 & 12.4 \\

DeepSeek-V4-Flash
& 89.4 & 74.2 & 15.1
& 70.2 & 49.3 & 20.8
& 66.0 & 54.0 & 12.0 \\

GPT-OSS-120B
& 78.6 & 29.4 & 49.2
& 52.7 & 17.6 & 35.1
& 28.0 & 13.3 & 14.7 \\

Grok-4.20
& 71.9 & 35.5 & 36.4
& 48.3 & 16.6 & 31.7
& 40.0 & 16.6 & 23.4 \\

Mistral-Large-3
& 64.5 & 23.6 & 40.9
& 32.4 & 10.3 & 22.1
& 21.1 & 6.9 & 14.3 \\

\midrule
\rowcolor{baseTaskColor!12}
\textbf{Average}
& 84.1 & 53.4 & 30.7
& 63.6 & 37.6 & 26.0
& 56.2 & 39.8 & 16.5 \\
\bottomrule
\end{tabular}
}

\caption{
Environmental novelty consistently degrades pass@1 (\%) across models and datasets.
Moreover, models with stronger base performance do not always exhibit smaller adaptation gaps.
Here, $\Delta_{\mathrm{adapt}}$ denotes the Base-Novel pass@1 gap in percentage points.
}
\label{tab:base-novel-performance}
\end{table}

%% file: sections/evaluating_adaptation.tex
\section{Evaluating Adaptation under Environmental Novelty}
\label{sec:evaluating-adaptation}

We now study adaptation in terminal agents using the three evaluation sets curated in Section~\ref{sec:agni}.

\textbf{Setup.}
We evaluate a diverse set of LLMs, including GPT-5.5, GPT-5.4-mini, Kimi-2.6, DeepSeek-V4-Flash, Grok-4.20, GPT-OSS-120B, and Mistral-Large-3.
Using a fixed Terminus-2 harness, we sample $K=5$ trajectories per model-task pair.
More details in Appendix~\ref{app:detailed-evaluation-agents}.

\textbf{Q5.1. Are Curated Novelties Relevant Across Models?}
Our base-novel pairs are constructed from assumptions derived from GPT-5.5 trajectories.
This raises the question of whether these novelties also affect the strategies used by other models.
We measure interact@1, the fraction of novel-task trajectories that interact with the changed component.
Across the three evaluation sets, interact@1 remains high (90.7--99.0\%), while successful trajectories that bypass the novelty are rare ($\leq 2.7\%$).
Thus, the curated novelties generally affect the strategies of other models rather than being specific to the trajectories used for construction.
More details in Appendix~\ref{app:cross-model-relevance}.

\textbf{Q5.2. How Does Novelty Affect Performance?}
Table~\ref{tab:base-novel-performance} reports pass@1 (\%) on the base and novel versions of all three datasets.
First, environmental novelty substantially reduces performance.
Average pass@1 drops from 84.1\% to 53.4\% on ET-eval, from 63.6\% to 37.6\% on TB-Lite, and from 56.2\% to 39.8\% on TB-2.
These correspond to adaptation gaps of 30.7, 26.0, and 16.5 percentage points, respectively.
Second, this degradation occurs across models and datasets, suggesting that it is not specific to any particular model or benchmark.
Third, strong base performance does not necessarily translate to better adaptation.
For instance, on ET-eval, GPT-5.4-mini achieves 94.5\% on base tasks, compared with 92.7\% for Kimi-2.6 and 89.4\% for DeepSeek-V4-Flash.
Under novelty, however, GPT-5.4-mini drops to 61.9\%, while Kimi-2.6 and DeepSeek-V4-Flash achieve 74.7\% and 74.2\%, respectively.
Likewise, GPT-OSS-120B outperforms Grok-4.20 on base tasks (78.6\% vs.\ 71.9\%) but falls below it on novel tasks (29.4\% vs.\ 35.5\%).
These ranking reversals persist even after normalizing performance degradation by base performance.
Overall, performance in the base environment can overestimate an agent's ability to adapt to environmental novelty.

\textbf{Q5.3. How Does Reasoning Effort Affect Adaptation?}
Adaptation requires recognizing environmental changes, diagnosing invalid assumptions, and revising strategy.
These adaptation stages may benefit from more test-time reasoning between interactions.
Table~\ref{tab:reasoning-effort} tests this hypothesis by reporting pass@1 (\%) on ET-eval across varying reasoning efforts.

\finishcompactwrap
\begin{wraptable}{r}{0.45\textwidth}
\input{sections/tables/agni-question-2}
\end{wraptable}
\fitcompactwrap
We observe that moving from low to high reasoning effort increases novel pass@1 by 20.4 percentage points for GPT-5.4-mini and 13.1 points for GPT-OSS-120B.
Their adaptation gaps also decrease, suggesting that greater reasoning effort improves adaptation for these models.
By contrast, for Kimi-2.6, medium effort yields the highest novel-environment performance, while low effort yields the smallest adaptation gap.
Thus, Kimi-2.6 shows a weaker, non-monotonic relationship between reasoning effort and adaptation.
This is consistent with prior findings that increased reasoning effort may not always improve performance and can reduce accuracy due to overthinking~\citep{zhou-etal-2026-thinking}.
Overall, reasoning effort often improves adaptation, but the gains can be model-dependent.

\finishcompactwrap
\begin{wrapfigure}{r}{0.45\textwidth}
\centering
    \includegraphics[
        width=\linewidth,
        trim=0 10pt 0 0,
        clip
    ]{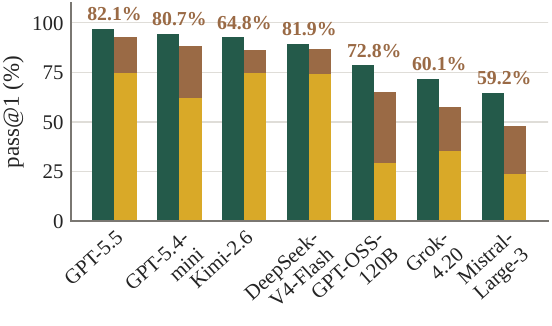}

    \caption{Novelty disclosure substantially recovers the adaptation gap on ET-eval. Bars show pass@1 for {\setlength{\fboxsep}{0.8pt}\colorbox{baseTaskColor}{\textcolor{white}{\textbf{base}}}} and {\setlength{\fboxsep}{0.8pt}\colorbox{novelTaskColor}{\textcolor{white}{\textbf{novel}}}} tasks, with
{\setlength{\fboxsep}{0.8pt}\colorbox{disclosedTaskColor}{\textcolor{white}{\textbf{disclosure gain} $\Delta_{\mathrm{disc}}$}}}; labels report corresponding recovery fraction $\rho_{\mathrm{rec}}$.}
    \label{fig:disclosure-recovery}
\end{wrapfigure}
\fitcompactwrap
\textbf{Q5.4. How Much of the Adaptation Gap Does Novelty Disclosure Recover?}
To measure how much of the adaptation gap can be recovered when the novelty is known in advance, we disclose the exact environmental change in the task instruction.
This allows the agent to plan for the changed environment directly, rather than encountering the change, diagnosing it, and revising a strategy built on its assumption.
Importantly, this setting does not measure adaptation itself, but how much of the adaptation gap is recoverable when the changed condition is known.
As shown in Figure~\ref{fig:disclosure-recovery}, disclosure improves performance across all evaluated models, recovering 71.7\% of the adaptation gap on average, with recovery fractions ranging from 59.2\% to 82.1\%.
However, degradation under undisclosed novelty does not necessarily mean that agents fail to observe the change.
As the next question shows, agents often observe the novelty but differ in how well they diagnose and respond to it.
Overall, explicit disclosure recovers much of the adaptation gap, and the remaining gap reflects the difficulty of acting successfully even when the novelty is known.

\finishcompactwrap
\begingroup
\setlength{\intextsep}{6pt}
\begin{figure}[H]
    \centering
    \begin{minipage}[t]{0.485\linewidth}
        \centering
        \includegraphics[width=0.94\linewidth]{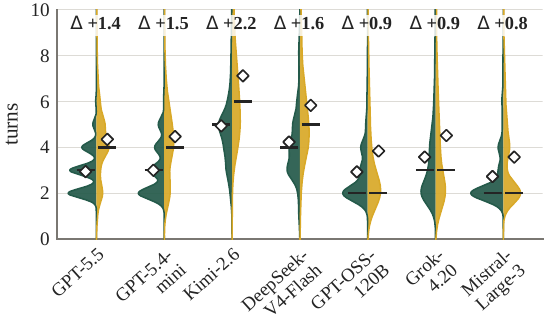}

        (a) Interaction length (\# turns)
    \end{minipage}
    \hfill
    \begin{minipage}[t]{0.485\linewidth}
        \centering
        \includegraphics[width=0.94\linewidth]{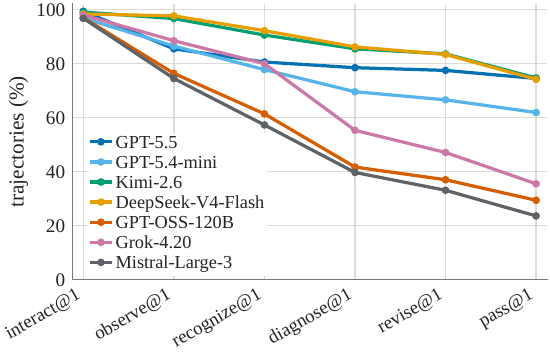}

        (b) Adaptation stages
    \end{minipage}
    \caption{
    Interaction behaviors associated with adaptation on ET-eval.
    (a) Better adapting models show larger increases in interaction length when going from {\setlength{\fboxsep}{0.8pt}\colorbox{baseTaskColor}{\textcolor{white}{\textbf{base}}}} to {\setlength{\fboxsep}{0.8pt}\colorbox{novelTaskColor}{\textcolor{white}{\textbf{novel}}}} tasks; lines mark medians, diamonds means, and labels report mean novel-base difference.
    (b) Better adapting models show more reliable progress from recognizing novelty to diagnosing it and revising their strategy.
    }
    \label{fig:interaction-behaviors}
\end{figure}
\endgroup

\textbf{Q5.5. Which Interaction Behaviors Are Associated with Better Adaptation?}
Novelty disclosure reveals how much performance is recoverable when the novelty is known, but not where adaptation fails when it is undisclosed.
We therefore compare the interaction behaviors of strongly and weakly adapting models.

To answer this, we first analyze interaction length, measured by the number of turns, across ET-eval base-novel pairs.
Figure~\ref{fig:interaction-behaviors}(a) shows that all models interact longer on novel tasks.
Stronger adapters (GPT-5.5, Kimi-2.6, and DeepSeek-V4-Flash) use 1.4--2.2 additional turns on average, compared with less than one additional turn for weaker adapters (GPT-OSS-120B, Grok-4.20, and Mistral-Large-3).
Thus, better adaptation is associated with sustaining longer interactions under novelty.

Second, we perform a turn-level analysis of ET-eval novel-task trajectories using a GPT-5.5-based judge.
At each turn, we annotate the agent's novelty-handling stage as \emph{interacted}, \emph{observed}, \emph{recognized}, \emph{diagnosed}, or \emph{revised}.
We also assign the agent's action a strategic intent, such as \emph{inspect}, \emph{diagnose}, \emph{retry}, or \emph{revise}.
Appendix~\ref{app:trajectory-annotations} provides details on the annotation process.
Figure~\ref{fig:interaction-behaviors}(b) reports the fraction of trajectories that reach each adaptation stage.
GPT-5.5 and Grok-4.20 recognize the novelty at similar rates (80.6\% vs.\ 80.0\%), but diverge sharply afterward: GPT-5.5 reaches diagnosis and strategy revision in 78.5\% and 77.5\% of trajectories, compared with 55.3\% and 47.1\% for Grok-4.20.
This suggests that the adaptation gap is not explained simply by whether agents recognize the novelty.
Successful adaptation also requires turning recognized novelty into a diagnosis and a revised strategy.
Appendix Tables~\ref{tab:retry-behavior} and~\ref{tab:behavioral-transitions} further show that weakly adapting models more often retry failed approaches, whereas stronger adapters more often transition from recognition to diagnosis and strategy revision.

\finishcompactwrap

%% file: sections/tables/agni-question-2.tex
\centering
\small
\setlength{\tabcolsep}{5pt}
\resizebox{\linewidth}{!}{%
\begin{tabular}{@{}llccc@{}}
\toprule
Model & Effort & Base & Novel & $\Delta_{\mathrm{adapt}}$ \\
\midrule
\multirow{3}{*}{GPT-5.4-mini}

& Low    & 85.4 & 41.5 & 44.0 \\
& Medium & 91.4 & 53.8 & 37.6 \\
& High   & 94.5 & 61.9 & 32.6 \\

\midrule

\multirow{3}{*}{GPT-OSS-120B}
& Low    & 75.0 & 22.2 & 52.8 \\
& Medium & 78.6 & 29.4 & 49.2 \\
& High   & 84.2 & 35.3 & 48.9 \\

\midrule

\multirow{4}{*}{Kimi-2.6}
& None   & 83.0 & 65.5 & 17.6 \\
& Low    & 90.1 & 72.7 & 17.4 \\
& Medium & 92.7 & 74.7 & 18.0 \\
& High   & 91.8 & 70.8 & 21.0 \\

\bottomrule
\end{tabular}
}

\caption{
Pass@1 (\%) on ET-eval across reasoning efforts.
Higher effort reduces $\Delta_{\mathrm{adapt}}$ for GPT-5.4-mini and GPT-OSS-120B but has a non-monotonic effect for Kimi-2.6.
}
\label{tab:reasoning-effort}

%% file: sections/post_training_adaptation.tex
\section{Improving Adaptation with Agentic Post-Training}
\label{sec:post-training-adaptation}

We now ask how adaptation changes during agentic post-training and whether simple changes to the training data and objective can improve it.

\textbf{Setup.}
We use Qwen3-8B (Qwen-8B) and Qwen3-14B (Qwen-14B) as our base models.
We curate training data from Endless Terminals by splitting 8,000 tasks into two disjoint 4,000-task sets, \(T_1\) and \(T_2\).
\textsc{AGNI} generates novel variants \(T_1'\) of \(T_1\).
This yields three equal-sized datasets:
\[
D_{\text{clean}} = T_1 \cup T_2,\qquad
D_{\text{novel-unpaired}} = T_2 \cup T_1',\qquad
D_{\text{novel-paired}} = T_1 \cup T_1'.
\]
The unpaired regime exposes the model to novelty across disjoint underlying tasks, whereas the paired regime presents the same tasks in both base and novel environments.
The paired construction, therefore, provides a contrast in the operating environment while keeping the underlying task fixed.
We evaluate on the ET-eval and TB-Lite base-novel pairs curated in Section~\ref{sec:agni}; all evaluation tasks are disjoint from the training tasks.
We select checkpoints using a separate set of 100 ET base tasks that are disjoint from both training and ET-eval.
We study GRPO~\citep{Shao2024DeepSeekMathPT} and ECHO~\citep{Shrivastava2026ECHOTA} as post-training approaches, with a primary focus on GRPO.
More details in Appendix~\ref{app:post-training-adaptation}.

\finishcompactwrap
\begin{wrapfigure}{r}{0.45\textwidth}
\centering
    \includegraphics[width=\linewidth]{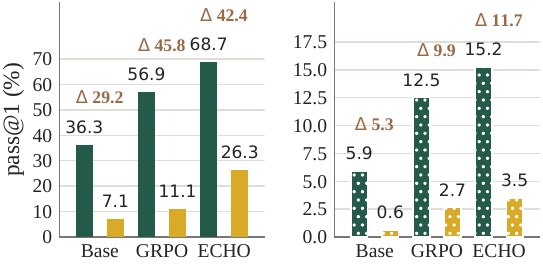}

    \caption{Generic post-training raises absolute performance but does not
    close the adaptation gap. Bars show
    {\setlength{\fboxsep}{0.8pt}\colorbox{baseTaskColor}{\textcolor{white}{\textbf{base}}}}
    and
    {\setlength{\fboxsep}{0.8pt}\colorbox{novelTaskColor}{\textcolor{white}{\textbf{novel}}}}
    pass@1; ET-eval is solid and TB-Lite dotted. $\Delta$ reports the adaptation gap.}
    \label{fig:posttraining-q1}
\end{wrapfigure}
\fitcompactwrap
\textbf{Q6.1. Does Generic Agentic Post-Training Improve Adaptation?}
We train Qwen-8B and Qwen-14B on \(D_{\text{clean}}\) using GRPO and ECHO and compare both post-trained models with the base model.
Figure~\ref{fig:posttraining-q1} shows that agentic post-training improves both base and novel performance, but does not narrow the adaptation gap.
On ET-eval, Qwen-8B Novel pass@1 increases from 7.1 to 11.1 with GRPO and 26.3 with ECHO, while Base pass@1 increases more sharply from 36.3 to 56.9 and 68.7, respectively.
Consequently, the adaptation gap remains large, increasing from 29.2 points to 45.8 with GRPO and 42.4 with ECHO.
We observe a similar trend in TB-Lite and with Qwen-14B (Appendix~\ref{app:post-training-adaptation}).

\finishcompactwrap
\begin{wrapfigure}{r}{0.45\textwidth}
\centering
    \includegraphics[width=\linewidth]{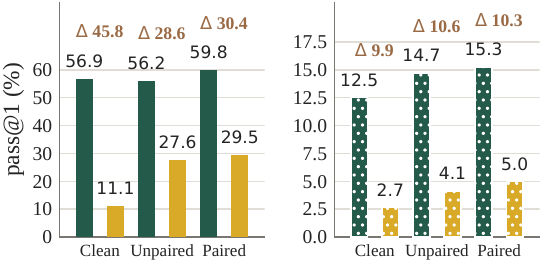}

    \caption{Novelty-augmented training narrows the ET-eval adaptation gap and
    transfers to TB-Lite. Bars show
    {\setlength{\fboxsep}{0.8pt}\colorbox{baseTaskColor}{\textcolor{white}{\textbf{base}}}}
    and
    {\setlength{\fboxsep}{0.8pt}\colorbox{novelTaskColor}{\textcolor{white}{\textbf{novel}}}}
    pass@1; ET-eval is solid and TB-Lite dotted. $\Delta$ reports the adaptation gap.}
    \label{fig:posttraining-q2}
\end{wrapfigure}
\fitcompactwrap
\textbf{Q6.2. Does Novelty-Augmented Training Improve Adaptation?}
We train Qwen-8B with GRPO on \(D_{\text{clean}}\), \(D_{\text{novel-unpaired}}\), and \(D_{\text{novel-paired}}\).
Figure~\ref{fig:posttraining-q2} shows that novelty-augmented training substantially improves held-out ET-eval novel performance while largely preserving base performance.
\(D_{\text{novel-unpaired}}\) increases novel pass@1 from 11.1 to 27.6 with similar base performance (56.9 vs.\ 56.2), reducing the adaptation gap from 45.8 to 28.6 points.
\(D_{\text{novel-paired}}\) further increases novel pass@1 to 29.5 and base pass@1 to 59.8, though its adaptation gap is slightly larger than unpaired training's (30.4 points).
The gain in novel pass@1 suggests that the paired contrast may help the model learn when environmental changes require a different strategy.
The gains also transfer to TB-Lite, where novel pass@1 increases from 2.7 under clean training to 4.1 and 5.0 with novel-unpaired and novel-paired training, respectively.
We observe a similar qualitative trend in Qwen-14B (Appendix~\ref{app:post-training-adaptation}).

\finishcompactwrap
\begin{wraptable}{r}{0.45\textwidth}
\centering
    \small
    \setlength{\tabcolsep}{3pt}
    \begin{tabularx}{\linewidth}{@{}>{\raggedright\arraybackslash}Xrrr@{}}
        \toprule
        \textbf{Method} & \textbf{Base} & \textbf{Novel} & \(\Delta_{\mathrm{adapt}}\) \\
        \midrule
        GRPO on \(D_{\text{novel-paired}}\) & 59.82 & 29.46 & 30.36 \\
        \rowcolor{baseTaskColor!12}
        + \textsc{Meta} Prompt & 61.25 & 32.59 & 28.66 \\
        \rowcolor{baseTaskColor!12}
        + \textsc{Meta} Prompt Training & 64.29 & 37.86 & 26.43 \\
        \midrule
        \rowcolor{disclosedTaskColor!12}
        + Recovery \& Verify & 69.55 & 38.00 & 31.55 \\
        \rowcolor{disclosedTaskColor!12}
        + Diagnosed Exploration & 53.93 & 25.14 & 28.79 \\
        \rowcolor{disclosedTaskColor!12}
        + Exploration \& Completion & 66.96 & 38.80 & 28.16 \\
        \bottomrule
    \end{tabularx}
    \caption{{\setlength{\fboxsep}{0.8pt}\colorbox{baseTaskColor!12}{Meta-action reasoning}} improves adaptation; {\setlength{\fboxsep}{0.8pt}\colorbox{disclosedTaskColor!12}{process rewards}} provide limited additional gains. Table reports ET-eval pass@1 (\%) and adaptation gap.}
    \label{tab:posttraining-q3}
\end{wraptable}
\fitcompactwrap
\textbf{Q6.3. Does Meta-Action Reasoning Improve Adaptation?}
Q5.5 (Section~\ref{sec:evaluating-adaptation}) shows that stronger adapters more reliably turn recognized novelties into diagnoses and strategy revisions.
Motivated by this, we ask whether explicitly reasoning about the intent behind each action, which we call the \emph{meta-action}, improves adaptation.
We modify the system prompt to generate one of five meta-actions alongside each action: \textsc{Execute}, \textsc{Inspect}, \textsc{Diagnose}, \textsc{Revise}, or \textsc{Verify}.
We refer to this as the \textsc{Meta} prompt; exact prompt changes are provided in Appendix~\ref{app:post-training-prompts}.
We restrict this experiment to paired training data, which performs relatively better on novel tasks.

We first apply the \textsc{Meta} prompt at inference time to the paired-data GRPO model, which raises held-out ET-eval novel pass@1 from 29.46\% to 32.59\%.
We then train a separate model from the same base checkpoint on paired data with the \textsc{Meta} prompt, keeping the reward purely outcome-based.
This further improves novel pass@1 to 37.86\% (Table~\ref{tab:posttraining-q3}).
Base pass@1 also increases from 59.82\% to 61.25\% and 64.29\%, respectively, though by a smaller magnitude.

We further explore adding simple process rewards over these meta-actions.
We consider three process level rewards: \textit{Recovery \& Verify}, \textit{Diagnosed Exploration}, and \textit{Exploration \& Completion} while holding all other training settings fixed.
Details of the design of these rewards are provided in Appendix~\ref{app:post-training-adaptation}.
\textit{Recovery \& Verify} and \textit{Exploration \& Completion} yield only small improvements in novel pass@1 over \textsc{Meta} training alone (38.00\% and 38.80\% vs.\ 37.86\%), while \textit{Diagnosed Exploration} reduces it to 25.14\%.
These rewards change the distribution of meta-actions (Appendix~\ref{app:reward-behavior}), but these changes translate into only limited improvements on novel tasks.

\finishcompactwrap

%% file: sections/discussion.tex
\section{Discussion and Limitations}
\label{sec:discussion-limitations}

As agents become more autonomous, users increasingly delegate not only task execution but also how those tasks are carried out.
This makes an agent's strategy and behavior as important as task success.
We study one aspect of this broader problem: whether an agent can reliably revise its strategy when environmental assumptions fail during execution.
Other aspects include resource efficiency~\citep{Garikaparthi2026CanLP,Foster2026AIRP} and safety~\citep{Vijayvargiya2025OpenAgentSafetyAC,li2026atbench}: a successful strategy may waste substantial time or compute, and task completion may involve harmful actions or violations of operational constraints.
Agent evaluation should therefore assess both task outcomes and behavior throughout execution.

\textsc{AGNI} curates task pairs by invalidating trajectory-relative assumptions while preserving task objectives and solvability.
A broader direction is to extend this pipeline to construct controlled settings for studying a wider range of agent behaviors.
For example, environments could test whether agents respect access boundaries around planted honeypots, consider alternative strategies under uncertainty, or abandon an unpromising strategy as contrary evidence accumulates.
This could enable controlled evaluation of how agents reason during execution, rather than evaluating only whether they succeed.

Our evaluation captures a bounded form of adaptation: recovery from curated environmental changes in terminal tasks.
It does not establish reliable adaptation under multiple changes, in very long-horizon tasks, or in open-ended deployment; measured failure rates only reflect performance on these simple curated tests.
LLM-assisted environment curation and evaluation may also introduce biases that our checks cannot fully eliminate.
Finally, computational costs restrict our post-training experiments to a limited set of models and training settings.
Despite these limitations, our results motivate environmental variation as a core dimension of agent training and evaluation.

%% file: sections/conclusion.tex
\section{Conclusion}
\label{sec:conclusion}

We study adaptation in terminal agents through environmental novelty.
To evaluate and train adaptation in controlled environments, we introduce \textsc{AGNI}, a scalable LLM-based pipeline to inject novelty.
\textsc{AGNI} invalidates assumptions underlying successful trajectories while preserving task objectives and solvability.
Across models and datasets, novelty substantially reduces performance, and strong base-task competence does not reliably predict adaptation.
Trajectory analysis shows that successful adaptation depends on progressing from recognizing a novelty to diagnosing it and revising the strategy.
Novelty-augmented post-training and explicit meta-action reasoning improve performance on novel tasks, while process rewards reshape the agent's behavior with only limited additional gains.
Our findings motivate treating environmental variation as a core dimension of agent training and evaluation, as well as evaluating agent strategy and behavior alongside task success.

%% file: sections/appendix/statements.tex
\section{AI Use, Ethics, and Reproducibility Statements}
\label{app:statements}

\subsection{AI Use Statement}
\label{app:ai-use}

Generative AI tools were used both as components of our experimental methodology and as auxiliary research tools.
GPT-5.5 was used within the \textsc{AGNI} pipeline for strategy analysis, novelty generation and implementation, behavioral validation, trajectory analysis, and realism evaluation, as described in Section~\ref{sec:agni} and the appendices.
Claude was used as an auxiliary tool for locating and summarizing potentially relevant literature, supporting code implementation and debugging, and polishing the grammar, clarity, and concision of author-written text.
Claude was not used as a source of research ideas, methodological decisions, interpretations, or claims.
All experimental designs, literature judgments, analyses, claims, and conclusions were reviewed and determined by the authors, who take full responsibility for the final manuscript.

\subsection{Ethics Statement}
\label{app:ethics}

This work evaluates agents on software benchmarks. LLM-generated outputs may
contain harmful content, and improved adaptation may also enable misuse.
Our evaluation measures task adaptation and does not establish the safety of
agent behavior or generated content. Generated environments and agent commands
should be executed with appropriate isolation and access controls.

\subsection{Reproducibility Statement}
\label{app:reproducibility}

We provide implementation and experimental details for environment generation,
agent evaluation, trajectory analysis, and post-training.
Algorithm~\ref{alg:agni} and Appendix~\ref{app:agni-details} describe the
\textsc{AGNI} pipeline, including model roles, generation budgets, environment
validation, and evaluation-set sampling. Appendix~\ref{app:agni-prompts}
provides the generation and analysis prompts and output schemas.
Appendices~\ref{app:detailed-evaluation-agents},
\ref{app:trajectory-annotations}, and~\ref{app:agni-realism} specify the
evaluation models and inference settings, evidence-based annotation criteria,
and realism-scoring protocol, respectively.

For post-training, Section~\ref{sec:post-training-adaptation} and
Appendix~\ref{app:post-training-adaptation} document the training-data
constructions, starting checkpoints, optimization hyperparameters, interaction
budgets, checkpoint-selection criterion, full agent prompts, and process-reward
definitions. We report model identifiers,
dated snapshots where available, and sampling settings to support replication
of the experimental setup. Exact numerical reproduction may nevertheless vary
because generation is stochastic and hosted model endpoints can change.

%% file: sections/appendix/detailed-related-work.tex
\section{Detailed Related Work}
\label{app:detailed-related-work}

\paragraph{Open-World Learning and Adaptation.}
Open-world learning studies how agents adapt when environmental novelty invalidates their assumptions~\citep{Langley2020OpenWorldLF,Boult2021TowardsAU,Kejriwal2024ChallengesEA}.
Prior work decomposes adaptation into detecting novelty, diagnosing its cause, and updating knowledge or behavior accordingly~\citep{Muhammad2021ANA,Mohan2023ADA,Goel2024ANC,Loyall2025ColtraneAD}.
Crucially, novelty is relative to the agent and its task: an environmental change may invalidate one strategy yet be irrelevant to another~\citep{Boult2021TowardsAU,Goel2024ANC}.
Much prior work studies embodied, simulated, or game-based environments, often using structured world models and dedicated components for novelty detection, diagnosis, and adaptations~\citep{Raibert2008BigDogTR,Kumar2021RMARM}.
In contrast, we study adaptation in general-purpose LLM agents operating through open-ended programmatic interfaces, where strategies and assumptions are largely implicit in interaction trajectories.

\paragraph{Robustness Evaluation in LLM Agents.}
Recent work studies the robustness of LLM agents by perturbing tool availability~\citep{trevino-etal-2025-benchmarking,wang2025hell,Liu2026PlanBenchXLEL,liu-etal-2026-costbench}, tool interfaces~\citep{Zhu2026WhenTF,Tian2026BeyondFC,wu2026can}, workspace contents~\citep{Li2026RepoMiragePR,Mahmud2026AJF,Englnder2026AgentsEB}, and runtime conditions~\citep{Jha2026AgentMT,Niu2026DuMateBenchEA}.
\textsc{AGNI} shares several design principles with some of these works, including paired base/modified tasks, undisclosed changes, preserved solvability, and perturbations to agent dependencies.
However, prior works largely predefine perturbation families, which are less suited for open-ended terminal tasks using heterogeneous resources and conditions~\citep{Ivison2026TmaxAS,merrill2026terminalbench}.
In terminal tasks, the relevant dependencies may depend on the agent's specific strategy rather than on the task alone.
Thus, \textsc{AGNI} starts from successful trajectories, identifies environmental assumptions that the agent relies on but are not guaranteed by the task instruction, and invalidates them while preserving solvability.
Moreover, since generated novelties are tied to a task-specific violated assumption, this enables a more controlled analysis of novelty adaptation than prior trajectory-level analyses of agent failure and recovery~\citep{Toh2026ScrambleToolBenchAS,kuang2026processlevel,Zhao2026FailureAA}.

\paragraph{Post-Training for Adaptive LLM Agents.}
Recent work post-trains agents for robustness to perturbations in tool-use~\citep{Vuddanti2025PALADINSL,zhang-etal-2026-robust,wu2026can,su-etal-2026-failure,Chen2026LearningTA,Hu2026SEALSC}, text-game~\citep{Ye2026LookBY}, and GUI environments~\citep{sun2026agenthijack}.
Our work differs along two axes.
First, what agents train on: prior work samples perturbations from predefined families, which are not suited for the heterogeneous action space of a terminal. \textsc{AGNI} instead constructs executable novel environments from assumptions revealed by the agent's successful trajectories.
Second, how agents are supervised: prior methods use task outcomes~\citep{Chen2026LearningTA,sun2026agenthijack}, teacher-written recovery demonstrations~\citep{Vuddanti2025PALADINSL,wu2026can}, externally supplied diagnoses~\citep{zhang-etal-2026-robust,Hu2026SEALSC}, or a single rewarded behavior such as reflection or exploration~\citep{su-etal-2026-failure,Ye2026LookBY}.
We instead reward the adaptation process: agents report actions with meta-actions such as inspecting, diagnosing, revising, and verifying, enabling process-level rewards for adaptive behaviors expressed through meta-actions.

\paragraph{Software Evolution and Resilience Testing.}
Prior work shows that software environments routinely change in ways that invalidate previously valid workflows.
Common examples include interface and dependency evolution, configuration errors, and changes in system state.
Interface and dependency evolution can break existing workflows: APIs evolve and break downstream clients~\citep{dig2006apis}, applications retain older dependency versions~\citep{kula2018dependencies}, and system policies and runtime defaults change over time~\citep{pep668,coreutils2026,busybox,autoconf2023,pep686}.
Configuration errors are prevalent across commercial and open-source systems and can cause difficult-to-diagnose failures~\citep{yin2011configuration,xu2015configuration}.
Changes in system state can expose failures that depend on particular combinations and orderings of otherwise ordinary events~\citep{yuan2014simpletesting}.
These kinds of changes motivate resilience testing and chaos engineering, which evaluate whether software systems continue to function reliably under perturbed conditions~\citep{basiri2016chaos,basiri2019automating}.
These works motivate both our study and our design of environmental novelty in terminal environments.
Analogous to resilience testing for software systems, we evaluate whether terminal agents remain reliable when the surrounding software environment changes.

%% file: sections/appendix/agni-details.tex
\input{sections/appendix/agni-details/a-pipeline-and-validation}

%% file: sections/appendix/agni-details/a-pipeline-and-validation.tex
\section{Details of the \textsc{AGNI} Pipeline}
\label{app:agni-details}

\input{sections/appendix/agni-details/a1-llm-roles}
\input{sections/appendix/agni-details/a2-base-strategy}
\input{sections/appendix/agni-details/a3-novelty-candidates}
\input{sections/appendix/agni-details/a4-novel-environments}
\input{sections/appendix/agni-details/a6-adaptive-behavior}
\input{sections/appendix/agni-details/b1-evaluation-set-curation}
\input{sections/appendix/agni-details/b3-representative-pairs}

%% file: sections/appendix/agni-details/a1-llm-roles.tex
\subsection{LLM Roles in \textsc{AGNI}}
\label{app:agni-models}

Table~\ref{tab:agni-model-roles} lists the roles in Algorithm~\ref{alg:agni};
Table~\ref{tab:agni-generation-config} gives their settings.
Appendix~\ref{app:agni-prompts} provides their prompts and output schemas.

\begin{table}[H]
    \centering
    \small
    \begin{tabular}{
        @{}
        p{0.20\linewidth}
        p{0.19\linewidth}
        p{0.53\linewidth}
        @{}
    }
        \toprule
        \textbf{Role} & \textbf{Symbol} & \textbf{Purpose} \\
        \midrule
        Solver
        & \(M_{\mathrm S}\)
        & Generates interaction trajectories in the base and novel environments. \\

        Strategy analyzer
        & \(M_{\mathrm A}\)
        & Summarizes observed task-solving behavior and extracts trajectory-grounded environmental assumptions. \\

        Novelty generator
        & \(M_{\mathrm N}^{\mathrm{dim}}, M_{\mathrm N}^{\mathrm{spec}}\)
        & Proposes environmental changes that target extracted assumptions while preserving the task objective. \\

        Implementer
        & \(M_{\mathrm N}^{\mathrm{impl}}\)
        & Instantiates a novelty specification as a modified environment \(E'\) and produces an adapted reference solution \(S'\). \\

        Behavioral observer
        & \(M_{\mathrm A}^{\mathrm{obs}}\)
        & Checks whether the injected novelty produces the expected change in behavior between the base and novel environments. \\

        Trajectory analyzer
        & \shortstack[l]{\textsc{Analyze}\\\textsc{Trajectories}}
        & Analyzes novel-task trajectories to characterize adaptation and identify unintended shortcuts or reward hacking. \\
        \bottomrule
    \end{tabular}
    \caption{LLM roles in \textsc{AGNI}.}
    \label{tab:agni-model-roles}
\end{table}

\begin{table}[H]
\centering
\small
\setlength{\tabcolsep}{4pt}
\renewcommand{\arraystretch}{1.08}
\begin{tabular}{@{}p{0.22\linewidth}p{0.72\linewidth}@{}}
\toprule
\textbf{Setting} & \textbf{Value} \\
\midrule
Model & GPT-5.5 (2026-04-24), all roles. \\
Reasoning effort & High for the solver; medium for other roles. \\
Sampling & Temperature and top-$p$ unset (API defaults); no fixed seed. \\
Output-token limits & 8{,}192 per solver turn; 32{,}000 per non-solver call. \\
Solver harness & Terminus-2 (Harbor~0.6.6); at most 32 turns. \\
Construction rollouts & \(K=8\) per base and novel task, across all three sources. \\
Candidate budgets
& \(N_{\mathrm{spec}}=5\) per applicable dimension;
up to \(N_{\mathrm{keep}}=8\) variants per base task. \\
Implementation repair & At most one round. \\
\bottomrule
\end{tabular}
\caption{\textsc{AGNI} model settings and construction budgets.}
\label{tab:agni-generation-config}
\end{table}

%% file: sections/appendix/agni-details/a2-base-strategy.tex
\subsection{Analyzing Base-Task Strategies}
\label{app:agni-characterize}

\paragraph{Base trajectories.}
For each base task \(\mathcal{T}_{\mathrm{base}}=(I,E,G)\), we sample
\(K\) independent trajectories from \(M_{\mathrm S}\) and retain the task only
if at least one passes the deterministic grader \(G\). This ensures nonzero
observed base performance: if the solver already scores zero, added difficulty
from novelty may leave the score unchanged, obscuring its effect.
The reference solution \(S\) provides a valid base strategy.

\paragraph{Strategy analysis.}
For each retained task, \(M_{\mathrm A}\) receives \((I,E,G)\), \(S\), and all
\(K\) trajectories with their pass/fail labels. It returns the task objective,
summaries of reference and solver behavior, and a comparison of successful and
failed trajectories. The summary records commands executed, tools and resources
used, files and paths accessed, and where successful and failed trajectories diverge.

\paragraph{Environmental assumptions.}
The analyzer extracts assumptions underlying the observed strategy. Both
assumptions and the novelties targeting them span four dimensions:

\begin{itemize}
    \item \textbf{Perception:} expected observations, including resource
    presence, absence, identity, and location.
    \item \textbf{Control:} how the solver can act, including available tools,
    commands, and interfaces.
    \item \textbf{Constraints:} permitted or feasible actions, including
    permissions, resource limits, and writable locations.
    \item \textbf{Dynamics:} action and environment behavior, including runtime
    behavior, dependency semantics, and the effects of commands or system operations.
\end{itemize}

\paragraph{Assumption schema.}
For each assumption, \(M_{\mathrm A}\) returns:
(i) its statement;
(ii) the sources relying on it (the reference solution, successful or failed
solver trajectories, or both);
(iii) supporting evidence;
(iv) why the assumption may be brittle; and
(v) what the solver would need to notice or do if it were invalidated.
These outputs ground candidate novelties in observed base-task behavior.
A subsequent dimension-identification call assigns the applicable categories.

The exact strategy-analysis prompt and output schema are provided in
Appendix~\ref{app:agni-prompts}.

%% file: sections/appendix/agni-details/a3-novelty-candidates.tex
\subsection{Synthesizing Novelty Descriptions}
\label{app:agni-synthesis}

\paragraph{Applicable novelty dimensions.}
Given the base task and strategy analysis from
Appendix~\ref{app:agni-characterize}, \(M_{\mathrm N}^{\mathrm{dim}}\)
selects dimensions in which minimal, task-preserving changes can invalidate
an extracted assumption and require different solver behavior.

\paragraph{Candidate generation.}
For each selected dimension \(d\), \(M_{\mathrm N}^{\mathrm{spec}}\)
receives \((I,E,G)\), the behavior summary, the associated assumptions, and
\(\mathcal{H}_d\): short descriptions of previously rejected candidates and
their failure reasons. This history is maintained separately for each base
task and updated after each rejection. The model generates
\(N_{\mathrm{spec}}=5\) candidate descriptions per dimension, each specifying
a novelty subtype, proposed environmental edits,
expected trajectory change, why adaptation is needed, why the task objective
is preserved, risks, and the two scores below.

\paragraph{Filtering and selection.}
We discard malformed descriptions and filter candidates using two scores
assigned by \(M_{\mathrm N}^{\mathrm{spec}}\), each on a 1--5 scale:
\emph{implementation plausibility} (1: impossible or too invasive;
3: requires moderate care; 5: trivial to implement) and
\emph{expected behavioral change} (1: identical solutions;
3: some visible change; 5: the original strategy fails).
An LLM compares short candidate summaries within each base task and removes
proposals for the same or very similar environmental changes.
We interleave the remaining descriptions across dimensions to give each
dimension equal opportunity in candidate selection.

Appendix~\ref{app:agni-prompts} provides the dimension-selection and
description-generation prompts and output schemas.

%% file: sections/appendix/agni-details/a4-novel-environments.tex
\subsection{Implementing and Validating Novel Tasks}
\label{app:agni-instantiation}

\paragraph{Implementation.}
For each candidate \(c\), \(M_{\mathrm N}^{\mathrm{impl}}\) receives
\((I,E,G)\), the reference solution \(S\), and the novelty description.
It produces a modified environment \(E'_c\) and an adapted reference solution
\(S'_c\), which provides one valid strategy for the modified task.

\paragraph{Task preservation.}
Edits are limited to those needed for the novelty, such as changes to environment
state, permissions, tools, dependencies, or runtime behavior. The task objective
and grader assertions must remain unchanged. \(G\) still denotes the same
success criterion. The agent must encounter the novelty through interaction,
without the instruction revealing it.

\paragraph{Implementation repair.}
Each candidate receives one implementation attempt. If building the environment
or executing \(S'_c\) fails, we return execution and grader feedback to
\(M_{\mathrm N}^{\mathrm{impl}}\) and allow one repair; candidates that still
fail are rejected. This repair step is omitted from Algorithm~\ref{alg:agni}.

\paragraph{Validation.}
Candidates must satisfy the task-preservation constraints and pass three checks
before behavioral validation (Appendix~\ref{app:agni-adaptation-validation}):
\begin{enumerate}
    \item \textbf{Task solvability:} the adapted reference succeeds,
    \(G(S'_c,E'_c)=1\).
    \item \textbf{Base-strategy invalidation:} the original reference fails,
    \(G(S,E'_c)=0\).
    \item \textbf{Solver recoverability:} among \(K\) fresh trajectories
    \(\mathcal{R}_c\) sampled from the construction solver \(M_{\mathrm S}\)
    in \(E'_c\), at least one passes \(G\).
\end{enumerate}

%% file: sections/appendix/agni-details/a6-adaptive-behavior.tex
\subsection{Validating Adaptive Behavior}
\label{app:agni-adaptation-validation}

After the executable checks in Appendix~\ref{app:agni-instantiation}, we verify
that successful solver behavior responds to the injected novelty.

\paragraph{Behavioral-effect validation.}
The observer \(M_{\mathrm A}^{\mathrm{obs}}\) receives the novelty description
and successful trajectories from the base and modified environments. It checks
for visible novelty and expected execution changes consistent with the targeted
assumption being invalidated. Candidates without evidence of the intended
behavioral effect are rejected.

\paragraph{Trajectory analysis.}
The trajectory analyzer examines all fresh solver trajectories
\(\mathcal{R}_c\) in \(E'_c\) step by step for adaptive recovery and reward
hacking. A valid recovery addresses the novelty and completes the original
task; it may differ from \(S'_c\). Reward hacking includes attempts to satisfy
checks by violating explicit task requirements, manipulating grader-relevant
artifacts, or replacing required operations without completing the task.

\paragraph{Acceptance.}
A candidate is retained only when the injected novelty produces the intended
behavioral effect, every grader-successful trajectory contains evidence of a
task-preserving adaptation, and no analyzed trajectory exhibits reward hacking.
An observed reward-hacking attempt causes rejection even if that trajectory
fails the grader.

Appendix~\ref{app:trajectory-annotations} describes the trajectory annotations;
Appendix~\ref{app:agni-prompts} provides the prompts and output schemas.

%% file: sections/appendix/agni-details/b1-evaluation-set-curation.tex
\subsection{Evaluation-Set Sampling}
\label{app:agni-eval-selection}

We sample the final 140 ET-eval, 118 TB-Lite, and 87 TB-2 base--novel pairs
using the same joint-stratification procedure for each source. The strata
capture task difficulty, novelty-induced difficulty, and realism.

\paragraph{Joint stratification.}
Each pair receives five tags (Table~\ref{tab:agni-eval-strata}). We assess
difficulty with a relatively strong sampler, GPT-5.4-mini, and a weaker one,
Ministral-14B. GPT-5.4-mini alone scores many ET-eval base tasks at 1.0,
leaving too little spread for difficulty stratification. The two samplers
estimate base-task difficulty, \(P_{\mathrm{base}}\), and relative
novelty-induced difficulty,
\(\delta=(P_{\mathrm{base}}-P_{\mathrm{novel}})/P_{\mathrm{base}}\).
The realism judge supplies occurrence plausibility, realization fidelity, and
mechanism family (Appendix~\ref{app:agni-realism}). We sample from the joint
cells formed by these tags, rather than balancing each tag separately. The
scores define strata; they are not exclusion thresholds.

\begin{table}[H]
    \centering
    \small
    \setlength{\tabcolsep}{5pt}
    \renewcommand{\arraystretch}{1.12}
    \caption{Strata used for joint sampling across ET-eval, TB-Lite, and TB-2.}
    \label{tab:agni-eval-strata}
    \begin{tabularx}{\linewidth}{@{}
        >{\raggedright\arraybackslash}p{0.35\linewidth}
        >{\raggedright\arraybackslash}X@{}}
        \toprule
        \textbf{Axis} & \textbf{Strata} \\
        \midrule
        Base-task difficulty
        & Easy, medium, hard \\
        Novelty-induced difficulty
        & Low, medium, high \\
        Occurrence plausibility
        & Low, medium, high \\
        Realization fidelity
        & Low, medium, high \\
        Mechanism family
        & Categorical labels from the frozen taxonomy \\
        \bottomrule
    \end{tabularx}
\end{table}

%% file: sections/appendix/agni-details/b3-representative-pairs.tex
\begingroup
\raggedbottom
\subsection{Examples from AGNI-Generated Base--Novel Task Pairs}
\label{app:agni-examples}

This section presents examples of novelties introduced by \textsc{AGNI} that
make agents' initial strategies insufficient while preserving the task objective
and grader. We identify the task source, describe the task and novelty, and
summarize the agents' responses in a short narrative and table. Comparisons
between adaptive and non-adaptive runs, which may come from the same model
or different models, illustrate adaptive behaviors.

\begin{table}[H]
\centering
\small
\setlength{\tabcolsep}{5pt}
\renewcommand{\arraystretch}{1.15}
\begin{tabularx}{\linewidth}{@{}p{0.18\linewidth}XX@{}}
\toprule
& \textbf{Adaptive behavior} & \textbf{Non-adaptive behavior} \\
\midrule
Agent & GPT-5.4-mini, \texttt{run\_02}. & GPT-5.4-mini, \texttt{run\_07}. \\
\addlinespace
Explanation & Reads the launch log and identifies \texttt{nohup} as the obstacle. & Leaves the cause unresolved and continues to rely on \texttt{nohup}. \\
\addlinespace
Action & Relaunches without \texttt{nohup}, using \texttt{python3 -m mlflow server \ldots\ \&}. & Retries launch commands that still use \texttt{nohup}. \\
\addlinespace
Check & Checks the server's health endpoint and confirms that it responds. & Launch attempts continue to fail; no working server is established. \\
\addlinespace
Result & Trains and registers the model; all three tests pass. & Does not complete the task; all three tests fail. \\
\bottomrule
\end{tabularx}
\caption{Adaptive and non-adaptive responses to a blocked launch command.}
\label{tab:agni-examples}
\end{table}

\paragraph{\textcolor{novelTaskColor!65!black}{Example 1: A blocked command for starting a server.}}\mbox{}

\noindent\textbf{Task Source:} TB-Lite [\texttt{mlflow-register}].

\noindent\textbf{Task.}
Start an MLflow tracking server in the background on port 8080, train a
linear regression model on random data with three input features, and
register the model as \texttt{gpt-5}.

\noindent\textbf{Novelty.}
Base runs start the server using \texttt{nohup}. The modified environment
adds a script at \texttt{/usr/local/bin/nohup} that is found before the
original command and rejects attempts to launch the server. The original
utility remains at \texttt{/usr/bin/nohup}, and starting the server directly
in the background still works. The task instructions and grader remain unchanged.

\noindent\textbf{Adaptive vs Non-Adaptive Run.}
Both GPT-5.4-mini runs encounter \texttt{Operation not permitted} when
starting the server (Table~\ref{tab:agni-examples}). The adaptive run reads
the launch log, identifies \texttt{nohup} as the obstacle, and starts the
server without it. It checks that the server responds, then trains and
registers the model. The non-adaptive run repeatedly uses \texttt{nohup}
despite the error and never establishes a working server. This comparison
suggests that adaptive behaviors such as inspecting errors, changing the
failed approach, and checking recovery can lead to success, whereas
non-adaptive behaviors such as repeating the failed approach can lead to failure.

\begin{table}[H]
\centering
\small
\setlength{\tabcolsep}{5pt}
\renewcommand{\arraystretch}{1.15}
\begin{tabularx}{\linewidth}{@{}p{0.18\linewidth}XX@{}}
\toprule
& \textbf{Adaptive behavior} & \textbf{Non-adaptive behavior} \\
\midrule
Agent & GPT-5.4-mini, \texttt{run\_00}. & GPT-5.5, \texttt{run\_02}. \\
\addlinespace
Explanation & Identifies the changed encoding as the reason direct searches miss the text. & Concludes that the logs contain none of the target words. \\
\addlinespace
Action & Uses Python to decode the logs as UTF-16 before counting matching lines. & Repeats \texttt{grep} searches without accounting for the encoding. \\
\addlinespace
Check & Confirms the UTF-16 encoding marker in the file bytes. & Accepts repeated zeros as evidence that the target words are absent. \\
\addlinespace
Result & Reports counts of 4, 3, and 8; all three tests pass. & Reports zero counts; the grader rejects the summary. \\
\bottomrule
\end{tabularx}
\caption{Adaptive and non-adaptive responses to changed log encoding.}
\label{tab:agni-example-log-encoding}
\end{table}

\paragraph{\textcolor{novelTaskColor!65!black}{Example 2: A changed text encoding in log files.}}\mbox{}

\noindent\textbf{Task Source:} TB-Lite [\texttt{log-summary}].

\noindent\textbf{Task.}
Read the log files in \texttt{/app/logs}, count lines containing
\texttt{ERROR}, \texttt{WARNING}, and \texttt{INFO}, and write the counts
in a CSV summary.

\noindent\textbf{Novelty.}
The modified environment stores the logs in UTF-16LE instead of UTF-8,
changing how the same text is stored as bytes. Extra zero bytes now separate
the letters in the target words, so the original searches return zero counts.
The filenames, log records, task instructions, and grader remain unchanged.

\noindent\textbf{Adaptive vs Non-Adaptive Run.}
Both runs initially report zero counts (Table~\ref{tab:agni-example-log-encoding}).
The adaptive GPT-5.4-mini run questions the result, inspects the file bytes
with Python, and identifies the UTF-16 encoding. It decodes the logs before
counting and produces the correct summary. The non-adaptive GPT-5.5 run also
notices suspicious zeros but repeats \texttt{grep} searches and concludes
that the target words are absent. This comparison suggests that adaptive
behaviors such as inspecting how data is stored and changing how it is read
can lead to success, whereas non-adaptive behaviors such as repeating an
unsuitable search and accepting unexplained zeros can lead to failure.

\begin{table}[H]
\centering
\small
\setlength{\tabcolsep}{5pt}
\renewcommand{\arraystretch}{1.15}
\begin{tabularx}{\linewidth}{@{}p{0.18\linewidth}X@{}}
\toprule
& \textbf{Adaptive behavior} \\
\midrule
Agent & GPT-5.4-mini, \texttt{run\_00}. \\
\addlinespace
Explanation & Suspects an incorrect program path in the first line or Windows line endings; does not directly inspect the extra character. \\
\addlinespace
Action & Reads the script and runs \texttt{bash /app/generate\_zk\_protocols.sh}, avoiding the broken first line. \\
\addlinespace
Check & Confirms that the script exists and has execution permission before changing how it is run. \\
\addlinespace
Result & Runs the generator and writes both reports; all 15 tests pass. \\
\bottomrule
\end{tabularx}
\caption{An adaptive response to a broken script launch.}
\label{tab:agni-example-script-shebang}
\end{table}

\paragraph{\textcolor{novelTaskColor!65!black}{Example 3: A script that fails to start despite being present.}}\mbox{}

\noindent\textbf{Task Source:} TB-Lite [\texttt{cryptographic-protocol-verifier}].

\noindent\textbf{Task.}
Run the provided script to generate cryptographic protocol transcripts,
analyze them, and write \texttt{verification\_results.json} and
\texttt{cryptographic\_analysis.txt}.

\noindent\textbf{Novelty.}
The script originally runs directly at \texttt{/app/generate\_zk\_protocols.sh}.
The modified environment adds a carriage-return character to its first line,
which specifies the program used to run it. The extra character becomes part
of that program's path, so direct execution fails even though the script
exists and has execution permission. The script body, task instructions,
and grader remain unchanged.

\noindent\textbf{Adaptive Run.}
This example follows one successful GPT-5.4-mini run
(Table~\ref{tab:agni-example-script-shebang}). The agent confirms that the
script exists and is executable but receives
\texttt{cannot execute: required file not found} when it tries to run it.
It suspects a problem with the first line or Windows line endings, reads the
script, and runs it explicitly through \texttt{bash}. The generator works,
and the agent completes both reports. This illustrates how adaptive behaviors
such as investigating an unexpected error and changing how a tool is called
can lead to success without repairing the tool itself.

\begin{table}[H]
\centering
\small
\setlength{\tabcolsep}{5pt}
\renewcommand{\arraystretch}{1.15}
\begin{tabularx}{\linewidth}{@{}p{0.18\linewidth}XX@{}}
\toprule
& \textbf{Adaptive behavior} & \textbf{Non-adaptive behavior} \\
\midrule
Agent & GPT-OSS-120B, \texttt{run\_00}. & GPT-OSS-120B, \texttt{run\_01}. \\
\addlinespace
Explanation & Finds that a configuration setting prevents pip from searching for packages. & Attributes the remaining failure to Python-version compatibility without confirming that explanation. \\
\addlinespace
Action & Inspects pip configuration and removes \texttt{no-index} from \texttt{/etc/pip.conf}. & Restores pip but leaves its global configuration unchecked. \\
\addlinespace
Check & Successfully installs \texttt{requests} after removing the setting. & Confirms pip's version but accepts \texttt{No matching distribution found} when testing installation. \\
\addlinespace
Result & Restores package installation; both tests pass. & Leaves installation broken and declares completion. Verifier setup also fails, so the task tests do not run. \\
\bottomrule
\end{tabularx}
\caption{Adaptive and non-adaptive responses to a package-installation restriction.}
\label{tab:agni-example-pip-configuration}
\end{table}

\paragraph{\textcolor{novelTaskColor!65!black}{Example 4: A setting that prevents pip from finding packages.}}\mbox{}

\noindent\textbf{Task Source:} TB-Lite [\texttt{broken-python}].

\noindent\textbf{Task.}
Repair the system-wide Python package installer, \texttt{pip}, so that it
can install packages.

\noindent\textbf{Novelty.}
The original task requires restoring missing packaging tools. The modified
environment also adds \texttt{no-index = true} to \texttt{/etc/pip.conf},
preventing pip from searching the package index for available packages.
Restoring pip alone therefore leaves package installation broken.
The task instructions and grader remain unchanged.

\noindent\textbf{Adaptive vs Non-Adaptive Run.}
Both GPT-OSS-120B runs restore pip but still encounter installation failures
(Table~\ref{tab:agni-example-pip-configuration}). The adaptive run inspects
pip's configuration, finds the restriction, removes it, and successfully
installs \texttt{requests}. The non-adaptive run confirms pip's version but
cannot install \texttt{requests}. It blames Python-version compatibility
without checking the configuration and declares the repair complete.
This comparison suggests that adaptive behaviors such as investigating a
remaining error and testing the required operation can lead to success,
whereas non-adaptive behaviors such as accepting an unchecked explanation
and stopping before the operation works can lead to failure.

\begin{table}[H]
\centering
\small
\setlength{\tabcolsep}{5pt}
\renewcommand{\arraystretch}{1.15}
\begin{tabularx}{\linewidth}{@{}p{0.18\linewidth}XX@{}}
\toprule
& \textbf{Adaptive behavior} & \textbf{Non-adaptive behavior} \\
\midrule
Agent & GPT-5.4-mini (high), \texttt{run\_01}. & GPT-5.4-mini (high), \texttt{run\_00}. \\
\addlinespace
Explanation & Suspects that an inherited \texttt{TAR\_OPTIONS} setting changes the stored names. & Recognizes the wrong names but does not identify the setting that changes them. \\
\addlinespace
Action & Unsets \texttt{TAR\_OPTIONS} when recreating and extracting the archive. & Repeats the archive command without changing the setting. \\
\addlinespace
Check & Lists the rebuilt archive and verifies the names \texttt{app.log} and \texttt{error.log}. & Sees the \texttt{raw/} prefix again but stops after extraction fails. \\
\addlinespace
Result & Extracts the log and completes the outputs; all four tests pass. & Leaves the archive incorrect and the extracted log missing; three tests fail. \\
\bottomrule
\end{tabularx}
\caption{Adaptive and non-adaptive responses to changed archive filenames.}
\label{tab:agni-example-tar-options}
\end{table}

\paragraph{\textcolor{novelTaskColor!65!black}{Example 5: A setting that changes filenames inside an archive.}}\mbox{}

\noindent\textbf{Task Source:} ET-eval [\texttt{task\_54fe38b3}].

\noindent\textbf{Task.}
Create \texttt{INC123\_logs.tar.gz} containing exactly \texttt{app.log} and
\texttt{error.log}, with no directory names before them. Also list the archive
contents, extract \texttt{error.log}, and write its SHA-256 checksum.

\noindent\textbf{Novelty.}
The original \texttt{tar} command stores the filenames as given. The modified
environment sets \texttt{TAR\_OPTIONS} to add \texttt{raw/} before each name.
The archive therefore contains \texttt{raw/app.log} and \texttt{raw/error.log},
so extracting \texttt{error.log} by its expected name fails.
The task instructions and grader remain unchanged.

\noindent\textbf{Adaptive vs Non-Adaptive Run.}
Both GPT-5.4-mini runs encounter \texttt{error.log: Not found in archive}
and see the unexpected \texttt{raw/} prefix when listing the contents
(Table~\ref{tab:agni-example-tar-options}). The adaptive run suspects an
inherited setting, recreates the archive with \texttt{TAR\_OPTIONS} unset,
checks the corrected names, and extracts the log. The non-adaptive run
recognizes the wrong names but repeats the same command and stops while
extraction still fails.
This comparison suggests that adaptive behaviors such as investigating tool
settings and verifying a correction can lead to success, whereas non-adaptive
behaviors such as repeating an unchanged command and stopping with an
unresolved error can lead to failure.

\begin{table}[H]
\centering
\small
\setlength{\tabcolsep}{5pt}
\renewcommand{\arraystretch}{1.15}
\begin{tabularx}{\linewidth}{@{}p{0.18\linewidth}XX@{}}
\toprule
& \textbf{Adaptive behavior} & \textbf{Non-adaptive behavior} \\
\midrule
Agent & GPT-OSS-120B, \texttt{run\_00}. & GPT-OSS-120B, \texttt{run\_02}. \\
\addlinespace
Explanation & Identifies that the script treated a repeated header as a release. & Blames a blank line at the end of the CSV for creating an extra file. \\
\addlinespace
Action & Reads the CSV, removes the generated files, and rebuilds the outputs while skipping header lines. & Does not read the CSV. Attempts to delete \texttt{release-.json}, although the extra file is named \texttt{release-version.json}. \\
\addlinespace
Check & Lists the regenerated files and confirms that exactly three remain. & Sees that the count is still four but declares the task complete. \\
\addlinespace
Result & Completes the task; all seven tests pass. & Leaves the extra file and log entry; two tests fail and five pass. \\
\bottomrule
\end{tabularx}
\caption{Adaptive and non-adaptive responses to a repeated CSV header.}
\label{tab:agni-example-repeated-header}
\end{table}

\paragraph{\textcolor{novelTaskColor!65!black}{Example 6: A repeated header in the input CSV.}}\mbox{}

\noindent\textbf{Task Source:} ET-eval [\texttt{task\_x1uj21}].

\noindent\textbf{Task.}
Read \texttt{releases.csv} and create exactly three JSON configuration files (manifests),
one for each software release, plus a three-line deployment log in input order.
Each file must be named \texttt{release-<version>.json}, using the version
from the input, and contain an integer port number.

\noindent\textbf{Novelty.}
The original CSV has one header listing the column names, followed by three
releases. The modified CSV repeats the header,
\texttt{version,app\_name,port}, before the third release. A script that skips
only the first line now treats the repeated header as a fourth release.
The task instructions and grader remain unchanged.

\noindent\textbf{Adaptive vs Non-Adaptive Run.}
Both GPT-OSS-120B runs finish their initial commands without an error but
report four files instead of three, including an extra
\texttt{release-version.json} (Table~\ref{tab:agni-example-repeated-header}).
The adaptive run reads the CSV, identifies the repeated header, and
regenerates the outputs while skipping header lines. It checks that exactly
three files remain and passes all tests. The non-adaptive run blames a blank
line, attempts to remove the wrong file, and stops with the count still at
four. This comparison suggests that adaptive behaviors such as investigating
the cause and verifying a correction can lead to success, whereas non-adaptive
behaviors such as acting on an unchecked explanation and stopping with an
unresolved error can lead to failure.

\begin{table}[H]
\centering
\small
\setlength{\tabcolsep}{5pt}
\renewcommand{\arraystretch}{1.15}
\begin{tabularx}{\linewidth}{@{}p{0.18\linewidth}XX@{}}
\toprule
& \textbf{Adaptive behavior} & \textbf{Non-adaptive behavior} \\
\midrule
Agent & GPT-5.5, \texttt{run\_01}. & GPT-5.5, \texttt{run\_03}. \\
\addlinespace
Explanation & Reads the startup error and identifies an environment setting that requests a missing database script. & Treats a successful build and the expected JAR path as sufficient to declare completion. \\
\addlinespace
Action & Sets the database connection explicitly in the application and rebuilds the JAR. & Configures the database through the properties file, which the environment setting overrides. \\
\addlinespace
Check & Launches the rebuilt JAR, confirms startup, and tests login. & Checks that the JAR exists, but never launches it or tests an endpoint. \\
\addlinespace
Result & Runs the service successfully; all ten tests pass. & The grader cannot start the service; all ten tests report setup errors. \\
\bottomrule
\end{tabularx}
\caption{Adaptive and non-adaptive responses to an overridden database setting.}
\label{tab:agni-example-datasource-precedence}
\end{table}

\paragraph{\textcolor{novelTaskColor!65!black}{Example 7: An environment setting overrides the application's database configuration.}}\mbox{}

\noindent\textbf{Task Source:} TB-Lite [\texttt{token-auth-websocket}].

\noindent\textbf{Task.}
Build a Spring Boot authentication service supporting login, token renewal,
one-time tokens, logout, and cleanup. Package it as an executable Java archive
(JAR) at \texttt{/app/build/libs/app.jar}. The grader launches this file and
tests the service on port 8080.

\noindent\textbf{Novelty.}
The base runs configure the database through \texttt{application.properties}.
The modified environment sets \texttt{SPRING\_DATASOURCE\_URL} to request
initialization from a missing file, \texttt{/tmp/missing-init.sql}.
This setting overrides the packaged properties: the application builds
successfully but fails when started. The task instructions and grader remain
unchanged.

\noindent\textbf{Adaptive vs Non-Adaptive Run.}
Both GPT-5.5 runs build the JAR, but only the adaptive run launches it
(Table~\ref{tab:agni-example-datasource-precedence}). It reads the missing-file
error, inspects the environment, and identifies the database override. It
sets the database connection explicitly in the application, rebuilds, and
checks startup and login. The non-adaptive run confirms that the JAR exists
and stops without launching it; the startup error appears only during
grading. This comparison suggests that adaptive behaviors such as testing
the application, diagnosing configuration conflicts, and verifying recovery
can lead to success, whereas non-adaptive behaviors such as stopping after
a successful build without testing runtime behavior can lead to failure.

\begin{table}[H]
\centering
\small
\setlength{\tabcolsep}{5pt}
\renewcommand{\arraystretch}{1.15}
\begin{tabularx}{\linewidth}{@{}p{0.18\linewidth}XX@{}}
\toprule
& \textbf{Adaptive behavior} & \textbf{Non-adaptive behavior} \\
\midrule
Agent & Kimi-2.6, \texttt{run\_03}. & Kimi-2.6, \texttt{run\_04}. \\
\addlinespace
Explanation & Recognizes that Git's saved decision has left the requested biography update missing. & Recognizes that Git kept the old biography, but treats the conflict as resolved. \\
\addlinespace
Action & Retrieves the updated biography from the project history and saves it. & Saves the result with the old biography still in place. \\
\addlinespace
Check & Reads the biography and checks whether it includes the requested Stanford update. & Reads the old biography but accepts Git's resolved status as sufficient. \\
\addlinespace
Result & Restores the updated biography and page layout; both tests pass. & Restores the layout but leaves the biography unchanged; one test fails and one passes. \\
\bottomrule
\end{tabularx}
\caption{Adaptive and non-adaptive responses to Git reusing an old decision.}
\label{tab:agni-example-rerere}
\end{table}

\paragraph{\textcolor{novelTaskColor!65!black}{Example 8: A saved Git decision keeps the old biography.}}\mbox{}

\noindent\textbf{Task Source:} TB-2 [\texttt{fix-git}].

\noindent\textbf{Task.}
Recover missing website edits: an updated biography mentioning Stanford and
a changed page layout. The edits remain in the project's Git history and
must be restored to its current version (\texttt{master}).

\noindent\textbf{Novelty.}
Normally, combining these versions makes Git flag conflicting edits and
wait for the agent to choose the intended text. The modified environment
enables Git's \texttt{rerere} feature to reuse an earlier decision:
keep the old biography. It also marks the file as ready to save.
Git still reports a merge error, but allows the agent to finalize the
result without restoring the updated biography. The instructions and grader
remain unchanged. The modified task also reconstructs the project's history,
preserving the required final contents and conflicting edits but simplifying earlier
versions and adding a named shortcut to the lost version.

\noindent\textbf{Adaptive vs Non-Adaptive Run.}
Both Kimi-2.6 runs find the missing edits, ask Git to combine them with the
current website, and see a notice that Git reused a previous decision
(Table~\ref{tab:agni-example-rerere}). Both read the biography and recognize
that Git kept the old text. The adaptive run notices that the Stanford update
is missing, retrieves the updated biography, and saves the corrected result.
The non-adaptive run treats Git's resolved status as sufficient, saves the
old biography, and declares the changes restored. This comparison suggests
that adaptive behaviors such as checking a tool's automatic decision against
the user's goal and correcting the result can lead to success, whereas
non-adaptive behaviors such as accepting the tool's status without ensuring
the requested changes are present can lead to failure.

\par
\endgroup

%% file: sections/appendix/agni-details/b4-realism-analysis.tex
\section{Realism Evaluation Details}
\label{app:agni-realism}
\label{app:realism}

We score two aspects of each validated novelty: \emph{occurrence plausibility},
whether the environmental condition could arise in practice, and
\emph{realization fidelity}, how faithfully its implementation reproduces that
condition (Table~\ref{tab:realism-rubric}). These scores do not affect acceptance
in Algorithm~\ref{alg:agni}; they are used afterward for characterization and
stratification into low (1--2), medium (3), and high (4--5) bins
(Appendix~\ref{app:agni-eval-selection}).

\begin{table}[H]
\centering
\small
\setlength{\tabcolsep}{5pt}
\begin{tabularx}{\linewidth}{@{}cXX@{}}
\toprule
\textbf{Score} & \textbf{Occurrence plausibility} & \textbf{Realization fidelity} \\
\midrule
1 & Implausible or specific to the evaluator & Unfaithful or incoherent implementation \\
2 & Technically possible but contrived & Crude proxy with major artificial artifacts \\
3 & Plausible in some deployments & Controlled abstraction preserving the central effect \\
4 & Recognizable real deployment or configuration change & Faithful simulation with minor simplifications \\
5 & Documented, observed, or widely recognized failure mode supported by the artifacts & Direct or near-exact reproduction of the real mechanism \\
\bottomrule
\end{tabularx}
\caption{The two realism scores are assigned separately using this rubric.}
\label{tab:realism-rubric}
\end{table}

\paragraph{Judging protocol.}
We use GPT-5.5 (2026-04-24) with medium reasoning effort, a 32{,}000-token initial
completion budget, and temperature unset. We retain one valid verdict per
novelty, with no score aggregation. Truncated or unparseable responses are
retried with a doubled token budget; missing required fields trigger one
corrective retry.

\paragraph{Inputs and evidence.}
The judge receives the agent-facing instruction, novelty specification,
environment diff, and changed files; reference solutions, tests, and grader
content are excluded. It identifies the underlying condition from the artifacts
and provides concrete evidence and rationale before each score. The prompt and output schema
are provided in Appendix~\ref{app:agni-prompts}.

\paragraph{Results.}
Table~\ref{tab:realism-results} reports the final evaluation sets. Before
stratification, mean plausibility/fidelity scores were 3.75/4.41 for ET
(283 candidate pairs), 3.81/4.51 for TB-Lite (370), and 3.74/4.35 for TB-2 (239).

\begin{table}[H]
\centering
\small
\setlength{\tabcolsep}{3pt}
\begin{tabular}{@{}lrcrrcrr@{}}
\toprule
& & \multicolumn{3}{c}{\textbf{Occurrence plausibility}}
& \multicolumn{3}{c}{\textbf{Realization fidelity}} \\
\cmidrule(lr){3-5}\cmidrule(l){6-8}
\textbf{Dataset} & \textbf{Pairs} & \textbf{Counts (1--5)} & \textbf{Mean} & \(\geq4\)
& \textbf{Counts (1--5)} & \textbf{Mean} & \(\geq4\) \\
\midrule
ET-eval & 140 & 0, 2, 43, 86, 9 & 3.73 & 68\% & 0, 0, 20, 35, 85 & 4.46 & 86\% \\
TB-Lite & 118 & 0, 6, 25, 76, 11 & 3.78 & 74\% & 0, 1, 11, 41, 65 & 4.44 & 90\% \\
TB-2 & 87 & 0, 2, 18, 60, 7 & 3.83 & 77\% & 0, 0, 12, 29, 46 & 4.39 & 86\% \\
\midrule
All & 345 & 0, 10, 86, 222, 27 & 3.77 & 72\% & 0, 1, 43, 105, 196 & 4.44 & 87\% \\
\bottomrule
\end{tabular}
\caption{LLM realism ratings for the final evaluation sets. Counts list scores
1--5; \(\geq4\) is the percentage rated 4 or 5, rounded to the nearest integer.
Median plausibility and fidelity are 4 and 5, respectively, in every set.}
\label{tab:realism-results}
\end{table}

%% file: sections/appendix/detailed-evaluation-results.tex
\section{Detailed Evaluation Results}
\label{app:detailed-evaluation-results}
\raggedbottom

The base, novel, and disclosed results in Section~\ref{sec:evaluating-adaptation}
use the same weighting over base--novel pairs.

\subsection{Evaluation Models and Inference Configuration}
\label{app:detailed-evaluation-agents}

Table~\ref{tab:evaluation-model-configs} lists the evaluated models and
reasoning settings; Table~\ref{tab:evaluation-inference-settings} gives the
shared inference configuration.

\begin{table}[H]
\centering
\small
\setlength{\tabcolsep}{4pt}
\begin{tabularx}{\linewidth}{@{}
>{\raggedright\arraybackslash}p{0.17\linewidth}
>{\raggedright\arraybackslash}p{0.11\linewidth}
>{\raggedright\arraybackslash}X
>{\raggedright\arraybackslash}p{0.14\linewidth}
>{\raggedright\arraybackslash}p{0.15\linewidth}@{}}
\toprule
\textbf{Model}
& \textbf{Provider}
& \textbf{Model / snapshot}
& \textbf{Main effort}
& \textbf{Other efforts} \\
\midrule
GPT-5.5
& OpenAI
& gpt-5.5 (2026-04-24)
& High
& -- \\

GPT-5.4-mini
& OpenAI
& gpt-5.4-mini (2026-03-17)
& High
& Medium, Low \\

GPT-OSS-120B
& OpenAI
& gpt-oss-120B
& Default
& High, Low \\

Kimi-2.6
& Moonshot
& Kimi-K2.6 (2026-04-20)
& Default
& High, Low, None \\

DeepSeek-V4-Flash
& DeepSeek
& DeepSeek-V4-Flash-0731 (2026-07-31)
& None
& -- \\

Grok-4.20
& xAI
& grok-4.20-beta-0309
& Non-reasoning
& -- \\

Mistral-Large-3
& Mistral
& Mistral-Large-3
& N/A
& -- \\
\bottomrule
\end{tabularx}

\caption{
Evaluation models. Default means \texttt{reasoning\_effort} was omitted
(approximately medium for GPT-OSS-120B and Kimi-2.6). Other efforts are used
in Table~\ref{tab:reasoning-effort}.
}
\label{tab:evaluation-model-configs}
\end{table}

\begin{table}[H]
\centering
\small
\setlength{\tabcolsep}{5pt}
\begin{tabularx}{\linewidth}{@{}lX@{}}
\toprule
\textbf{Setting} & \textbf{Value} \\
\midrule
Agent harness & Terminus-2, at most 32 interaction turns. \\
Rollouts & \(K=5\) per model--task pair, unless otherwise specified. \\
Completion budget & 8{,}192 tokens per model response. \\
Sampling & Temperature and top-$p$ unset (deployment defaults); no explicit seed. \\
Verifier timeout & \(10\times\) the configured verifier timeout. \\
\bottomrule
\end{tabularx}
\caption{Shared settings for the main evaluation.}
\label{tab:evaluation-inference-settings}
\end{table}

\subsection{Cross-Model Relevance of \textsc{AGNI} Novelties}
\label{app:cross-model-relevance}

For each novelty, GPT-5.5 identifies affected tools, paths, and command patterns, which we match
against executed commands. Table~\ref{tab:cross-model-relevance} reports
interact@1 for base and novel
rollouts: the fraction interacting with the affected component, averaged as
for pass@1. For novel rollouts, bypass@1 is the fraction of successful rollouts
without interaction; Fail w/o Int.\ is the fraction of failed rollouts without
interaction.

\newcommand{\AgniInteractionTableWidth}{1.00\textwidth}

\begin{table}[H]
\centering
\resizebox{\AgniInteractionTableWidth}{!}{%
\begin{tabular}{l|cccc|cccc|cccc}
\toprule
& \multicolumn{4}{c|}{ET-eval}
& \multicolumn{4}{c|}{TB-Lite}
& \multicolumn{4}{c}{TB-2} \\

& \multicolumn{2}{c}{Interact@1}
& \multicolumn{1}{c}{}
& \multicolumn{1}{c|}{}
& \multicolumn{2}{c}{Interact@1}
& \multicolumn{1}{c}{}
& \multicolumn{1}{c|}{}
& \multicolumn{2}{c}{Interact@1}
& \multicolumn{1}{c}{}
& \multicolumn{1}{c}{} \\

Model
& Base & Novel & Bypass@1 & Fail w/o Int.
& Base & Novel & Bypass@1 & Fail w/o Int.
& Base & Novel & Bypass@1 & Fail w/o Int. \\
\midrule

GPT-5.5
& 99.1 & 99.0 & 0.7 & 1.2
& 96.9 & 96.3 & 1.7 & 7.2
& 96.6 & 98.3 & 1.4 & 1.4 \\

GPT-5.4-mini
& 95.3 & 95.9 & 2.7 & 3.8
& 92.7 & 94.9 & 1.5 & 8.6
& 91.2 & 95.0 & 0.7 & 8.1 \\

Kimi-2.6
& 97.3 & 98.4 & 1.6 & 0.0
& 94.2 & 97.8 & 0.8 & 3.3
& 92.6 & 96.1 & 2.3 & 4.2 \\

DeepSeek-V4-Flash
& 96.1 & 97.8 & 1.3 & 3.4
& 95.4 & 96.6 & 0.7 & 5.4
& 94.6 & 95.5 & 0.8 & 4.2 \\

GPT-OSS-120B
& 95.3 & 95.3 & 1.7 & 4.3
& 91.0 & 93.2 & 0.3 & 7.8
& 93.8 & 94.5 & 0.0 & 6.4 \\

Grok-4.20
& 95.4 & 97.0 & 1.7 & 2.0
& 91.3 & 90.7 & 1.0 & 10.0
& 93.1 & 95.2 & 0.0 & 5.8 \\

Mistral-Large-3
& 94.5 & 96.0 & 1.3 & 3.6
& 91.7 & 91.4 & 1.0 & 8.5
& 92.4 & 93.3 & 0.0 & 7.2 \\

\bottomrule
\end{tabular}
}

\caption{Cross-model relevance of curated novelties (\%).}
\label{tab:cross-model-relevance}
\end{table}

\subsection{Retry Prevalence and Persistence}
\label{app:retry-analysis}

\emph{Retry prevalence} measures trajectories with at least one Retry step;
\emph{retry persistence} measures
$P(\mathrm{Retry}_{t+1}\mid\mathrm{Retry}_t)$ over transitions after the
novelty first becomes observable. Retry means repeating a failed strategy or a
superficial variant that preserves its assumption
(Appendix~\ref{app:trajectory-annotations}).

\begin{table}[H]
\centering
\small
\setlength{\tabcolsep}{10pt}
\begin{tabular}{lcc}
\toprule
& \textbf{Retry prevalence} & \textbf{Retry persistence} \\
\textbf{Model}
& Rollouts with $\geq 1$ Retry (\%)
& $P(\mathrm{Retry}_{t+1}\mid\mathrm{Retry}_t)$ (\%) \\
\midrule
GPT-5.5             & 9.8  & 9  \\
GPT-5.4-mini        & 15.2 & 50 \\
Kimi-2.6            & 18.1 & 6  \\
DeepSeek-V4-Flash   & 17.4 & 10 \\
GPT-OSS-120B        & 27.8 & 40 \\
Grok-4.20           & 40.7 & 45 \\
Mistral-Large-3     & 28.1 & 35 \\
\bottomrule
\end{tabular}
\caption{Retry prevalence and persistence on ET-eval novel rollouts (\%).}
\label{tab:retry-behavior}
\end{table}

\subsection{Adaptation Stage Definitions}
\label{app:adaptation-stage-definitions}

\begin{table}[H]
\centering
\small
\begin{tabular}{p{0.15\linewidth} p{0.78\linewidth}}
\toprule
\textbf{Stage} & \textbf{Definition} \\
\midrule

\textbf{Interacted}
& An executed command touches an affected tool, path, or file. Detected deterministically; no visible effect is required. \\

\textbf{Observed}
& Terminal output exposes novelty-related evidence, such as an error, unexpected result, or changed file state. Recognition is not required. \\

\textbf{Recognized}
& The agent explicitly identifies the observation as unexpected or inconsistent with its assumptions, without necessarily identifying the cause. \\

\textbf{Diagnosed}
& The agent identifies the changed condition sufficiently to explain the problem and guide a response. Restating a symptom is insufficient. \\

\textbf{Revised}
& A changed strategy successfully removes or bypasses the obstacle while preserving the task. Failed attempts and exploits do not count. \\

\textbf{Solved}
& The agent completes the original task under novelty using a task-preserving strategy. \\

\bottomrule
\end{tabular}
\caption{Stages of novelty handling used in trajectory analysis.}
\label{tab:adaptation-stage-definitions}
\end{table}

\subsection{Behavioral Transitions and Conditional Success}
\label{app:behavioral-transitions}

\begin{table}[H]
\centering
\small
\setlength{\tabcolsep}{3pt}
\begin{tabularx}{\linewidth}{@{}>{\raggedright\arraybackslash}Xcccccc@{}}
\toprule
\textbf{Model}
& \textbf{Diag.$\rightarrow$Rev.}
& \textbf{Diag.$\rightarrow$Retry}
& \shortstack{\textbf{Premature}\\\textbf{Done}}
& \shortstack{\textbf{Pass $|$}\\\textbf{Diag.}}
& \shortstack{\textbf{Pass $|$}\\\textbf{Rev.+Diag.}}
& \shortstack{\textbf{Pass $|$}\\\textbf{Rev. w/o Diag.}} \\
\midrule
GPT-5.5             & 71 & 5  & 14 & 96 & 96 & 68 \\
Kimi-2.6            & 41 & 4  & 11 & 86 & 87 & 50 \\
DeepSeek-V4-Flash   & 47 & 5  & 12 & 86 & 86 & 70 \\
GPT-5.4-mini        & 57 & 11 & 20 & 87 & 86 & 59 \\
GPT-OSS-120B        & 30 & 10 & 21 & 65 & 64 & 22 \\
Grok-4.20           & 31 & 14 & 16 & 61 & 64 & 20 \\
Mistral-Large-3     & 27 & 17 & 31 & 57 & 57 & 16 \\
\bottomrule
\end{tabularx}
\caption{
Detailed behavioral transitions and conditional success rates on ET-eval.
\textbf{Diag.$\rightarrow$Rev.} and \textbf{Diag.$\rightarrow$Retry} measure
revision or retry immediately after diagnosis.
\textbf{Premature Done} denotes trajectories that finalize immediately after
a failed or partial step.
The final columns report success conditioned on diagnosis and on whether
strategy revision occurs with or without a preceding diagnosis.
All values are percentages.
}
\label{tab:behavioral-transitions}
\end{table}

For every model, revision preceded by diagnosis is associated with higher
success than revision without diagnosis. Table~\ref{tab:behavioral-transitions}
also shows how often models retry or stop after a failed step, and how often
diagnosis leads to task success.

%% file: sections/appendix/agni-details/b2-trajectory-annotations.tex
\section{Analyzing Trajectories for Adaptive Behavior}
\label{app:trajectory-annotations}

We use a GPT-5.5-based analyzer to label each step of a novel-task trajectory.
It receives the task instruction, novelty specification, grader outcome, and
full interaction. These labels support the final checks in
Algorithm~\ref{alg:agni}: every grader-successful trajectory must show
task-preserving adaptation, and no trajectory may show reward hacking. They
also locate where unsuccessful trajectories stop progressing toward recovery.

\paragraph{Strategic intents.}
A step can have multiple substantive intents. The analyzer assigns:

\begin{itemize}
    \setlength{\topsep}{0pt}
    \setlength{\itemsep}{0pt}
    \setlength{\parsep}{0pt}
    \item \textbf{Inspect:} examining task inputs, files, or requirements before
    attempting a solution.

    \item \textbf{Execute:} making the first attempt that produces task output.

    \item \textbf{Diagnose:} investigating the cause of a visible failure or
    anomaly.

    \item \textbf{Explore:} finding an alternative method without applying it yet.

    \item \textbf{Retry:} repeating a failed strategy or a superficial variant
    that retains the failed assumption.

    \item \textbf{Revise:} applying a task-preserving fix for the novelty.

    \item \textbf{Repair:} finishing required artifacts after the novelty is handled.

    \item \textbf{Verify:} checking outputs against the task requirements.

    \item \textbf{Finalize:} requesting or confirming completion without further
    substantive changes.

    \item \textbf{Exploit:} passing visible checks while violating task requirements.

    \item \textbf{Other:} a substantive intent not captured by the labels above.
\end{itemize}

\paragraph{Adaptation stages.}
Separately, the analyzer marks when a step establishes each stage:

\begin{itemize}
    \setlength{\topsep}{0pt}
    \setlength{\itemsep}{0pt}
    \setlength{\parsep}{0pt}
    \item \textbf{Observed:} an observation exposes evidence specific to the change.

    \item \textbf{Recognized:} the agent treats that evidence as unexpected or
    inconsistent with its strategy.

    \item \textbf{Diagnosed:} the agent identifies the changed condition well
    enough to fix it.

    \item \textbf{Revised:} a changed method clears the novelty obstacle;
    the task may still be unfinished.

    \item \textbf{Solved:} the agent completes or verifies the original task
    under novelty with a task-preserving method.
\end{itemize}

We count a stage as reached only when the trajectory supports it. Seeing an
error does not establish recognition or diagnosis; trying a different command
does not establish a successful revision. The grader determines task success.
\emph{Solved} records evidence of how the task was completed and does not
replace the grader result.

\paragraph{Evidence grounding.}
Labels must be supported by the agent's visible reasoning, commands, observed
outputs, or deterministic command effects. The novelty specification and grader
outcome provide context but cannot establish an intermediate stage: a passing
grader result alone does not show that the agent recognized or diagnosed the
change. A generic error also does not count as \emph{Observed} unless it
specifically points to the changed condition. We count \emph{Recognized} or
\emph{Diagnosed} only when the agent engages with the evidence, rather than
merely receiving it in terminal output.

\paragraph{Retention predicates.}
For retention, adaptation must change the method that relied on the invalidated
assumption and still satisfy the original task. Repeating a failed method or
changing only an unrelated detail does not qualify. An \emph{Exploit} label
flags a substitute evaluator-facing artifact or a violation of the required
task operation. A different method that genuinely completes the task remains
valid even if it differs from \(S'_c\).

\paragraph{Verification-dependent novelties.}
Some changes yield a clean command result but incorrect task artifacts. We mark
these as requiring an explicit check of the resulting state to expose the
novelty, distinguishing failure to uncover evidence from failure to act on
evidence already seen. The trajectory analyses count progress through these
stages, including observation, diagnosis, revision, and task completion. The
exact prompt and output schema appear in Appendix~\ref{app:agni-prompts}.

%% file: sections/appendix/post-training-details.tex
\section{Improving Adaptation with Agentic Post-Training}
\label{app:post-training-adaptation}

We use the training hyperparameters of ECHO~\citep{Shrivastava2026ECHOTA}
in our post-training experiments (Table~\ref{tab:post-training-hyperparameters}).
The data, prompts, and rewards vary as described below. We train 8B models
for at most 500 steps and 14B models for at most 300 steps.

\begin{table}[H]
\centering
\small
\begin{tabularx}{\linewidth}{@{}lX@{}}
\toprule
\textbf{Setting} & \textbf{Value used in our experiments} \\
\midrule
Optimizer & AdamW; $\beta_1=0.9$, $\beta_2=0.95$ \\
Weight decay & $0.01$ \\
Learning rate & $10^{-6}$, constant; no warmup \\
Gradient clipping & $0.2$ \\
Batch size & $16$ \\
Rollouts per prompt & $16$ \\
GRPO clipping & $\epsilon_{\mathrm{lo}}=0.2$, $\epsilon_{\mathrm{hi}}=0.28$ \\
KL penalty & None \\
Advantage normalization & Within each prompt group \\
Loss aggregation & Per sequence \\
Training temperature & $0.8$ \\
Training interaction limit & $16$ turns \\
Training context limit & $16$k tokens \\
Generation limit & $2{,}048$ tokens per turn \\
Training timeouts & Agent: $600$\,s; verifier: $120$\,s \\
ECHO auxiliary weight & $\lambda=0.05$ (ECHO runs only) \\
ECHO prediction targets & Terminal output, excluding harness warnings;
normalized by total observation length \\
\bottomrule
\end{tabularx}
\caption{Training hyperparameters used in our experiments, adopted from
\citet{Shrivastava2026ECHOTA}. ECHO-specific settings apply only to ECHO runs;
run lengths are stated above.}
\label{tab:post-training-hyperparameters}
\end{table}

\subsection{Starting Checkpoints}

All runs start from \texttt{Qwen/Qwen3-8B} or \texttt{Qwen/Qwen3-14B},
respectively~\citep{Yang2025Qwen3TR}. Within each model size, the starting
checkpoint is fixed across comparisons.

\paragraph{Checkpoint selection.}
We select checkpoints using performance on 100 base tasks derived from ET that
are disjoint from both ET-eval and all training splits. ET-eval therefore remains
disjoint from both training and checkpoint selection. TB-Lite provides evaluation
on a separate task source.

\subsection{Controlled Training Comparisons}

For Qwen3-8B, we compare the clean, novelty-unpaired, and novelty-paired
training sets defined in Section~\ref{sec:post-training-adaptation}.
Table~\ref{tab:post-training-14b-paired} reports the corresponding clean and
novelty-paired GRPO comparison for Qwen3-14B.

\begin{table}[H]
\centering
\small
\begin{tabular}{lcccc}
\toprule
& \multicolumn{2}{c}{\textbf{ET-eval}} & \multicolumn{2}{c}{\textbf{TB-Lite}} \\
\cmidrule(lr){2-3}\cmidrule(l){4-5}
\textbf{Qwen3-14B GRPO} & \textbf{Base} & \textbf{Novel} & \textbf{Base} & \textbf{Novel} \\
\midrule
Clean & 65.3 & 24.6 & 17.2 & 2.9 \\
Novel-paired & 67.1 & 35.5 & 18.9 & 5.5 \\
\bottomrule
\end{tabular}
\caption{Qwen3-14B GRPO pass@1 (\%) on ET-eval and TB-Lite.}
\label{tab:post-training-14b-paired}
\end{table}

\subsection{META Interface and Process-Reward Comparisons}
\label{app:post-training-meta}

\paragraph{Prompt and action interface.}
The META prompt adds a required \texttt{meta\_action} field to every bash call,
with five possible values:
\textsc{Execute}, \textsc{Inspect}, \textsc{Diagnose}, \textsc{Revise}, and
\textsc{Verify}. It also tells the agent when to use each value based on its
interaction history. The complete prompts appear in
Appendix~\ref{app:post-training-prompts}.

\paragraph{Controlled comparisons.}
All Q6.3 runs use Qwen3-8B and \(D_{\text{novel-paired}}=T_1\cup T_1'\).
We evaluate the paired-data GRPO checkpoint with and without
the META prompt, keeping its weights fixed. We train the META model from the
base checkpoint with the META prompt and binary task reward. The three
process-reward arms also start from the base checkpoint and use the same data,
prompt, and otherwise matched training settings; they differ in reward and are
not successive training stages.

\paragraph{Process-reward arms.}
Each arm adds a small shaping term to the binary task reward: \(R=r+S\), where
\(r=1\) only when the verifier passes. Shaping uses the recorded command tags,
commands, exit codes, and \texttt{done} call. Missing or invalid tags incur one
0.1 penalty per rollout. Unless noted, event credit is paid in full on a pass
and at 25\% on a failure. A call fails when its exit code is nonzero; a mutating
call changes the environment.

\textit{Recovery \& Verify} gives 0.1 credit when the same
command succeeds after an earlier failure and a \textsc{Diagnose} or \textsc{Revise} tag, and 0.1 for a
read-only \textsc{Verify} call after the last environment change. Each credit
is awarded at most once.

\textit{Diagnosed Exploration} gives 0.1 for each of up to two
distinct new mutating fixes after the first failure followed by \textsc{Diagnose}
or \textsc{Revise}; repeating the failed command earns no credit. It subtracts
0.05 if any command fails and 0.05 for each repeat of the same command without
an intervening environment change, up to three.
Thus, a failed rollout cannot profit from creating failures to earn fix credit.

\textit{Exploration \& Completion} uses the same exploration
reward and penalties, then adds 0.2 for read-only verification after the last
change and 0.1 for calling \texttt{done}. These last two credits require task
success. The rewards use observable command patterns, not a judgment that the
agent's diagnosis was correct.

\begin{figure}[H]
    \centering
    \includegraphics[width=0.75\linewidth]{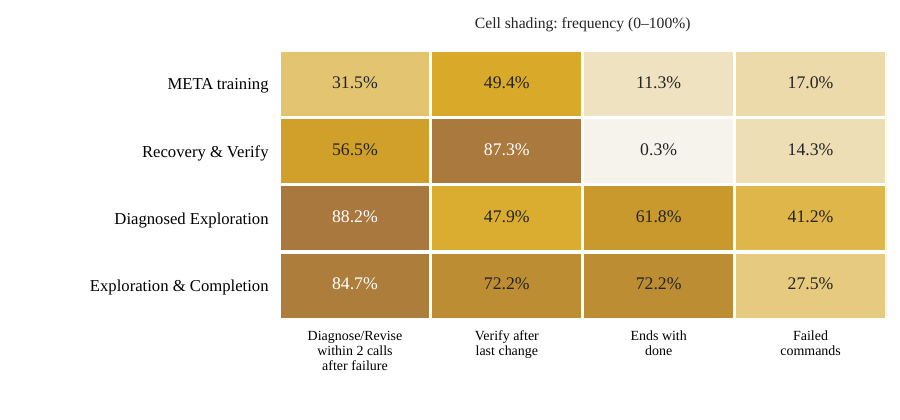}
    \caption{\textbf{Process rewards change behavior but yield limited gains on novel tasks.}
    Held-out ET-eval, 1,120 novel rollouts per arm. Failed commands are a
    percentage of calls; Table~\ref{tab:posttraining-q3} reports task success.}
    \label{fig:posttraining-q3-behavior}
\end{figure}

\paragraph{Held-out evaluation.}
We evaluate on 140 held-out ET-eval base--novel pairs with eight novel rollouts per pair.
Base scores use the corresponding paired-task weighting
(Table~\ref{tab:posttraining-q3}). Figure~\ref{fig:posttraining-q3-behavior}
uses all 1,120 novel rollouts per arm to measure diagnosis or revision within
two calls after a failure, verification after the last change, explicit
\texttt{done} calls, and the fraction of calls that fail. These measures record
actions, not whether the diagnosis or final answer was correct.

\paragraph{Behavioral effects of process rewards.}
\label{app:reward-behavior}
Relative to binary-reward META training, \textit{Recovery \& Verify} raises
verification after the last change from 49.4\% to 87.3\%, but only 0.3\% of
rollouts call \texttt{done}. \textit{Diagnosed Exploration} raises diagnosis
or revision within two calls after failure from 31.5\% to 88.2\%, while failed
calls rise from 17.0\% to 41.2\%. \textit{Exploration \& Completion} reaches
84.7\% on diagnosis or revision and 72.2\% on both verification and completion;
novel pass@1 rises only from 37.86\% to 38.80\%.

\subsection{Normal and META Prompts}
\label{app:post-training-prompts}

The listings reproduce both full prompts.
\textcolor{baseTaskColor}{Teal text} marks additions in the META prompt.

\begingroup
\makeatletter
\renewcommand{\thetcb@cnt@promptbox}{\thesection.\arabic{tcb@cnt@promptbox}}
\makeatother
\lstset{moredelim=**[is][\color{baseTaskColor}]{<<META>>}{<<ENDMETA>>}}
\begin{promptbox}[prompt:post-training-normal]{Normal Prompt (non-META)}
# Tools

You have access to the following functions:

<tools>
{"type": "function", "function": {"name": "bash", "description": "Execute a bash command in the terminal.", "parameters": {"type": "object", "properties": {"command": {"type": "string", "description": "The bash command to execute"}, "timeout": {"type": "integer", "description": "Timeout in seconds (default 30)"}}, "required": ["command"]}}}
{"type": "function", "function": {"name": "done", "description": "Mark the task as complete. Call this when you have verified the task is solved.", "parameters": {"type": "object", "properties": {}}}}
</tools>

If you choose to call a function ONLY reply in the following format with NO suffix:

<tool_call>
<function=example_function_name>
<parameter=example_parameter_1>
value_1
</parameter>
<parameter=example_parameter_2>
This is the value for the second parameter
that can span
multiple lines
</parameter>
</function>
</tool_call>

<IMPORTANT>
Reminder:
- Function calls MUST follow the specified format: an inner <function=...></function> block must be nested within <tool_call></tool_call> XML tags
- Required parameters MUST be specified
- You may provide optional reasoning for your function call in natural language BEFORE the function call, but NOT after
- If there is no function call available, answer the question like normal with your current knowledge and do not tell the user about function calls
</IMPORTANT>

You are a highly capable Linux terminal agent. Complete the user's task by running commands and verifying the result. When the task is complete, call done.
\end{promptbox}

\begin{promptbox}[prompt:post-training-meta]{META Prompt}
# Tools

You have access to the following functions:

<tools>
{"type": "function", "function": {"name": "bash", "description": "Execute a bash command in the terminal.", "parameters": {"type": "object", "properties": {<<META>>"meta_action": {"type": "string", "enum": ["EXECUTE", "INSPECT", "DIAGNOSE", "REVISE", "VERIFY"], "description": "Meta-strategic goal of this command. Details described in <IMPORTANT> tags."}, <<ENDMETA>>"command": {"type": "string", "description": "The bash command to execute"}, "timeout": {"type": "integer", "description": "Timeout in seconds (default 30)"}}, "required": [<<META>>"meta_action", <<ENDMETA>>"command"]}}}
{"type": "function", "function": {"name": "done", "description": "Mark the task as complete. Call this when you have verified the task is solved.", "parameters": {"type": "object", "properties": {}}}}
</tools>

If you choose to call a function ONLY reply in the following format with NO suffix:

<tool_call>
<function=example_function_name>
<parameter=example_parameter_1>
value_1
</parameter>
<parameter=example_parameter_2>
This is the value for the second parameter
that can span
multiple lines
</parameter>
</function>
</tool_call>

<IMPORTANT>
Reminder:
- Function calls MUST follow the specified format: an inner <function=...></function> block must be nested within <tool_call></tool_call> XML tags
- Required parameters MUST be specified
- You may provide optional reasoning for your function call in natural language BEFORE the function call, but NOT after
- If there is no function call available, answer the question like normal with your current knowledge and do not tell the user about function calls
<<META>>- In each turn involving a bash call, first think/reason (content in <think> ... </think> tags) which meta-action is appropriate given the interaction history, then think/reason about the specific command to take conditioned on that meta-action
- Meta-action is the strategic purpose of the next action: EXECUTE, INSPECT, DIAGNOSE, REVISE, or VERIFY; choose it according to the primary purpose of the command.
- The `meta_action` parameter in the subsequent `bash` function call MUST match the meta-action chosen in your reasoning/thinking
- Use meta-actions strategically based on the current situation: continue executing when the approach is making progress; when observations reveal missing information, unexpected behavior, failed assumptions, or that the approach is no longer suitable, inspect, diagnose, or revise as appropriate before proceeding; verify important results when needed.
- Use EXECUTE when you already know what approach to use and the next command directly carries it out
- Use INSPECT when you need additional information about the environment or current state before deciding what action to take
- Use DIAGNOSE when a command failed, an observation was unexpected, or an assumption appears false, and you need to determine the cause
- Use REVISE when evidence shows that the approach being used is blocked, invalid, or unlikely to succeed, and the next command begins carrying out a different approach
- Use VERIFY when checking whether a previous action succeeded or whether an intermediate or final task requirement has been satisfied<<ENDMETA>>
</IMPORTANT>

You are a highly capable Linux terminal agent. Complete the user's task by running commands and verifying the result. When the task is complete, call done.
\end{promptbox}
\endgroup

\subsection{Qwen3-14B Generic Post-Training Results}
\label{app:post-training-14b}

\begin{figure}[H]
\centering
\includegraphics[width=0.75\linewidth]{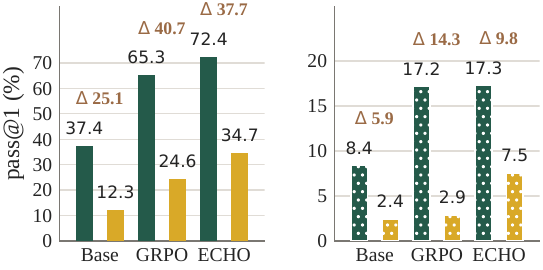}
\caption{Qwen3-14B generic post-training on ET-eval (solid) and TB-Lite
(dotted). Bars show {\setlength{\fboxsep}{0.8pt}\colorbox{baseTaskColor}{\textcolor{white}{\textbf{base}}}}
and {\setlength{\fboxsep}{0.8pt}\colorbox{novelTaskColor}{\textcolor{white}{\textbf{novel}}}}
pass@1 (\%); \(\Delta\) marks the adaptation gap.}
\label{fig:post-training-14b}
\end{figure}

Figure~\ref{fig:post-training-14b} reports the 14B clean-data comparison from
Q6.1. GRPO and ECHO improve base and novel pass@1, but the adaptation gap
widens under both methods.

%% file: sections/appendix/agni-details/c-prompts-and-output-schemas.tex
\section{Model Prompts Used in \textsc{AGNI}}
\label{app:agni-prompts}

We provide the prompts and structured output schemas used by each LLM-based
component of \textsc{AGNI}.
Each model call serves a distinct role in the pipeline. The inputs shown below
are the complete prompt-level inputs for that call.
The prompt text and JSON schemas are shown together so that each model call can
be reproduced without consulting a separate schema. Line wrapping is adjusted
only for typesetting; template placeholders are enclosed in braces.

Table~\ref{tab:agni-prompt-index} maps each prompt to its role and corresponding
operation in Algorithm~\ref{alg:agni}.
The shared definition in Prompt~\ref{prompt:novelty-definition} is interpolated
wherever the placeholder \texttt{\{novelty\_definition\}} appears.

\begin{table}[ht]
    \centering
    \small
    \begin{tabular}{@{}p{0.10\linewidth}p{0.46\linewidth}p{0.37\linewidth}@{}}
        \toprule
        \textbf{Prompt} & \textbf{Purpose} & \textbf{Usage in Algorithm~\ref{alg:agni}} \\
        \midrule
        \ref{prompt:novelty-definition} & Defines novelty for reuse across prompts & Lines 3, 4, 6, 11, 16: shared model context \\
        \ref{prompt:behavior-summary} & Characterizes the base strategy and assumptions & Line 3: \(M_{\mathrm A}\) \\
        \ref{prompt:dimension-proposal} & Identifies applicable novelty dimensions & Line 4: \(M_{\mathrm N}^{\mathrm{dim}}\) \\
        \ref{prompt:novelty-specification} & Synthesizes novelty specifications & Line 6: \(M_{\mathrm N}^{\mathrm{spec}}\) \\
        \ref{prompt:novelty-implementation} & Instantiates \(E'_c\) and adapted \(S'_c\) & Line 11: \(M_{\mathrm N}^{\mathrm{impl}}\) \\
        \ref{prompt:novelty-repair} & Repairs a failed task instantiation & Line 11: \(M_{\mathrm N}^{\mathrm{impl}}\) retry (implicit) \\
        \ref{prompt:solution-delta} & Validates the expected trajectory change & Lines 16--17: \(M_{\mathrm A}^{\mathrm{obs}}\) \\
        \ref{prompt:trajectory-analysis} & Analyzes intent, outcome, and adaptation progress & Lines 18--20: \textsc{AnalyzeTrajectories} \\
        \ref{prompt:disclose-novelty} & Discloses the change without suggesting a solution & After Algorithm~\ref{alg:agni}: constructs the disclosed-novelty condition \\
        \ref{prompt:realism-judgment} & Scores condition plausibility and realization fidelity & After Algorithm~\ref{alg:agni}: audits novelty realism \\
        \bottomrule
    \end{tabular}
    \caption{Prompt-to-operation mapping for the \textsc{AGNI} pipeline and disclosed-novelty evaluation. Line numbers refer to Algorithm~\ref{alg:agni}.}
    \label{tab:agni-prompt-index}
\end{table}

\input{sections/appendix/agni-details/prompts/p01-shared-novelty-definition}
\input{sections/appendix/agni-details/prompts/p02-base-strategy-characterizer}
\input{sections/appendix/agni-details/prompts/p03-novelty-dimension-identifier}
\input{sections/appendix/agni-details/prompts/p04-novelty-candidate-synthesizer}
\input{sections/appendix/agni-details/prompts/p05-novel-task-instantiator}
\input{sections/appendix/agni-details/prompts/p06-novel-task-instantiation-repair}
\input{sections/appendix/agni-details/prompts/p07-adaptive-behavior-validator}
\input{sections/appendix/agni-details/prompts/p08-trajectory-adaptation-analyzer}
\input{sections/appendix/agni-details/prompts/p09-novelty-disclosure-generator}
\input{sections/appendix/agni-details/prompts/p10-novelty-realism-judge}

%% file: sections/appendix/agni-details/prompts/p01-shared-novelty-definition.tex
\FloatBarrier
\begin{promptbox}[prompt:novelty-definition]{Shared Novelty Definition}
(*@\promptheading{Definition}@*)
An environmental novelty is a task-preserving change to the execution environment that invalidates an assumption underlying a previously successful strategy. On task T, the agent's strategy succeeds in environment E. In the novel task T', the objective and success criterion remain unchanged, but environment E' invalidates an assumption used by that strategy. The agent must discover the change through interaction and adapt, for example by exploring, replanning, switching tools, resolving permissions, or handling inputs more robustly. If the original behavior still succeeds without meaningful revision, the change is not a strong test of adaptation.

(*@\promptheading{Task Definition}@*)
A task T consists of:
* instruction: what the agent must accomplish;
* environment: its files, tools, permissions, dependencies, and runtime state;
* evaluator: the tests or grading harness that determine success; and
* reference solution: a known successful strategy for T.

(*@\promptheading{Criteria for Valid Novel Task}@*)
A novel task T'=(I,E',G) is valid only if all of the following hold:
1. Task-objective preservation: T' has the same underlying objective as T=(I,E,G); only the execution environment changes.
2. Success-criterion preservation: the grader represents the same notion of success. Mechanical updates for a relocated path or changed invocation must not alter which behavior or final state counts as successful.
3. Environmental change: the novelty changes an operating condition in E and invalidates an assumption used by a successful trajectory; it does not change the task objective.
4. Task solvability: at least one adapted strategy can achieve the original objective in E'.
5. Base-strategy invalidation: behavior that relied on the targeted assumption is insufficient in E', producing an interpretable behavioral difference between T and T'.
6. Robustness and non-trivial adaptation: the change should measurably reduce pass@1 for a weaker or less-careful agent relative to T by breaking a default assumption on which it reflexively relies. Recovery must require the agent to inspect the environment, diagnose the violated assumption, or revise its strategy. Reject changes that are irrelevant, trivially discoverable, trivially bypassed, or leave successful behavior effectively unchanged.
7. Autonomous discovery: the change is not announced, described, or hinted at in the instruction, READMEs, comments, or other task-facing artifacts. The agent must discover it by inspecting state, executing commands, or interpreting errors and outputs. Modify the instruction only when necessary to keep the original objective coherent, never to reveal the novelty.

(*@\promptheading{Constraint}@*)
Invalidate a trajectory-relevant assumption rather than introducing a change that is immediately discoverable and mechanically resolved.
* Weak novelty: relocate a file or rename a directory when one obvious inspection step reveals the new location. This does not meaningfully disrupt the strategy.
* Strong novelty: invalidate an assumption during execution and require revision after new evidence. Examples include a missing or shadowed tool with a non-obvious alternative; silent tool-behavior drift; a permission barrier encountered mid-task; a misleading decoy; or a runtime, version, locale, or interface change that alters semantics or output.

(*@\promptheading{Novelty Dimensions}@*)
Assign the novelty to one broad dimension:
* perception: what information or state is present and where it appears, including file layout, filenames, decoys, clutter, hidden state, and stale artifacts;
* control: how the agent acts, including shell, tool, CLI, or API substitutions, command syntax, and utility availability;
* constraints: what the agent is allowed or able to do, including permissions, non-root access, write restrictions, time, memory, disk, and network access; or
* dynamics: how actions behave, including package or runtime versions, defaults, ordering, locale, timezone, and parser or runtime behavior.

(*@\promptheading{Optional Modifier Tags}@*)
Use these only as tags, not as broad novelty dimensions:
* instruction_underspecified: the instruction provides fewer explicit hints while preserving the objective;
* residual_state: stale, temporary, or cached artifacts are already present; and
* idempotence_cleanup: repeated execution or cleanup of extra artifacts matters.

(*@\promptheading{Required Novelty Description}@*)
Every proposed novelty must state:
* broken_assumption: what the original solution or successful agent assumed;
* required_adaptation: what the agent must notice or do differently; and
* expected_failure: how a non-adaptive agent fails.

(*@\promptheading{Identifying a Consequential Novelty}@*)
Find an assumption used by successful trajectories, such as a clean initial state, exact file format, available tool, default permissions, or fixed path. It must not be required by the evaluator. Break only that assumption, keep the evaluator's substantive assertions unchanged, and force a different solution path already allowed by those assertions.

(*@\promptheading{Example Patterns}@*)
These patterns are examples, not a checklist. Use diverse mechanisms within each dimension rather than repeatedly proposing the same changes. Every novelty must preserve the task objective and substantive success criterion.
* Pre-existing state: a required file or directory already exists with incorrect contents, type, or permissions. A naive mkdir, touch, append, or overwrite fails or produces the wrong result, requiring inspection and repair through chmod, truncate, replacement, or cleanup.
* Decoy resources: a sibling file, directory, executable, or artifact resembles the target. Globbing, broad copying, or selecting the most salient match acts on the decoy; the agent must identify the correct resource.
* Format perturbation: comments, indentation, leading whitespace, a header row, or another formatting convention changes structure without changing semantics. Fixed-column or exact-text parsing fails, requiring robust parsing or inspection.
* Tool removal or shadowing: remove, shadow, or alter a reflexively used tool, such as tail, grep, sha256sum, date, or a gawk feature. The agent must diagnose the failure and use an alternative such as Python, sed, awk, or openssl.
* Tool discovery: the expected tool is unavailable but an equivalent exists; several similar tools exist but only one is appropriate; or the correct command or version must be inferred from --version, configuration, documentation, or other evidence.
* Permission or umask change: permissions, ownership, privilege level, or the default umask invalidates assumed write access, executability, or permissive defaults. The agent must inspect and correct the constraint.
* Renamed or undiscovered input: move or rename a hard-coded input. The agent must find it through ls, globbing, metadata, or context. Prefer cases requiring interpretation rather than one obvious lookup.
* Stale state or required cleanup: extra configuration lines, temporary files, locks, stale outputs, or partially completed state interfere with the base strategy and must be reconciled or removed.
* Semantic drift: the interface appears unchanged, but locale, timezone, encoding, sort order, runtime, or dependency version changes its behavior. The command still runs but produces a subtly incorrect result that must be recognized and addressed.
* Forced recovery: a natural first action from the successful strategy fails or produces an informative mismatch. The agent must interpret the error, output, or log and recover through diagnosis and replanning.

\end{promptbox}

%% file: sections/appendix/agni-details/prompts/p02-base-strategy-characterizer.tex
\FloatBarrier
\begin{promptbox}[prompt:behavior-summary]{Base Strategy Characterizer}
(*@\promptheading{Role}@*)
You are analyzing a terminal task T and solver trajectories collected on T.

(*@\promptheading{Objective}@*)
Characterize the successful base strategy and identify the environmental assumptions it relies on. Produce a compact, evidence-grounded summary that will support later novelty generation. Do not propose novel tasks or environmental changes in this step.

(*@\promptheading{Definition of Novelty}@*)
{novelty_definition}

(*@\promptheading{Inputs}@*)
* Original task files:
{task_files}
* Reference solution (solve.sh):
{reference}
* Base-task solver trajectories with pass/fail labels:
{trajectories}

(*@\promptheading{Requirements}@*)
* Return ONLY valid JSON.
* Treat the tests and reference solution as a evidence of the task objective.
* Use only evidence supported by the task files, reference solution, or trajectories.
* Use exact commands/file paths only when important.
* Separate observed behavior from inferred assumptions.
* If something is unclear, say "unclear".
* If no solver trajectory passed, set fields about successful trajectories to "" or [].
* Summarize behavior for a reviewer; do not reproduce exhaustive logs.

(*@\promptheading{Output Format}@*)
Return exactly one JSON object with this schema:
{
  "task_tldr":
    "<1-3 sentences: objective the tests actually check>",

  "solution_summary": {
    "reference":
      "<how solve.sh achieves the objective>",
    "frontier_successes":
      "<how successful agent runs achieved it; '' if none passed>",
    "frontier_failures":
      "<common reasons unsuccessful attempts failed; '' if none failed>"
  },

  "pass_fail_contrast": {
    "what_successful_agents_did":
      "<main behaviors that led to success>",
    "what_failed_agents_missed":
      "<main behaviors/assumptions/errors behind failures>",
    "key_difference":
      "<clearest behavioral difference between pass and fail>"
  },

  "important_commands_and_files": {
    "commands": [
      "<important recurring commands only>"
    ],
    "files_or_paths": [
      "<important files/paths inspected, modified, or assumed>"
    ]
  },

  "observed_assumptions": [
    {
      "assumption":
        "<path/tool/shell/version/permission/network/cwd/etc. assumption>",
      "who_relied_on_it":
        "reference | successful_agents | failed_agents | both",
      "evidence":
        "<brief concrete evidence>",
      "why_it_may_be_brittle":
        "<why changing it could require adaptation>",
      "how_can_assumption_be_broken_for_novelty":
        "<what an agent would need to notice or change if this assumption broke>"
    }
  ]
}
\end{promptbox}
\FloatBarrier

%% file: sections/appendix/agni-details/prompts/p03-novelty-dimension-identifier.tex
\FloatBarrier
\begin{promptbox}[prompt:dimension-proposal]{Novelty Dimension Identifier}
(*@\promptheading{Objective}@*)
Given a terminal task T and its behavior summary, identify the broad novelty dimensions that admit a task-preserving, implementable novelty for this task and are likely to produce an observable change in the agent's trajectory.

(*@\promptheading{Shared Novelty Definition}@*)
{novelty_definition}

(*@\promptheading{Inputs}@*)
* Original task files:
{task_files}
* Base Strategy Characterizer output:
{summary}

(*@\promptheading{Requirements}@*)
* Return ONLY valid JSON.
* Include only dimensions that can be implemented for this task with minimal changes.
* Do not include a dimension merely because it is generally possible.
* Ground each dimension in an observed assumption from the behavior summary.
* If no good dimension exists, return {"applicable": []}.
* Do not propose concrete file edits yet.

(*@\promptheading{Output Format}@*)
Return exactly one JSON object with this schema:
{
  "applicable": [
    {
      "dimension":
        "perception | control | constraints | dynamics",
      "why_this_task":
        "<why this dimension can bite this task specifically>",
      "broken_assumption":
        "<assumption from reference/agents that could be broken>",
      "expected_adaptation":
        "<what the novelty agent would need to notice or do differently>",
      "risk":
        "<main reason this dimension may fail or become non-task-preserving>"
    }
  ]
}
\end{promptbox}

\FloatBarrier

%% file: sections/appendix/agni-details/prompts/p04-novelty-candidate-synthesizer.tex
\begin{promptbox}[prompt:novelty-specification]{Novelty Candidate Synthesizer}
(*@\promptheading{Objective}@*)
Propose {n} distinct, concrete, task-preserving novelty specifications for terminal task T. Every specification must belong to the broad novelty dimension {dimension}.

Each specification is a plan, not an implementation. Prefer minimal, auditable changes that can be implemented and are grounded in how agents actually solved T.

(*@\promptheading{Constraints}@*)
* Do not add irrelevant noise.
* Do not change the task objective.
* Do not reveal the answer or the environmental change to the agent.
* Do not require many unrelated edits.

(*@\promptheading{Shared Novelty Definition}@*)
{novelty_definition}

(*@\promptheading{Inputs}@*)
* Past failure context:
{failure_context}
* Base Strategy Characterizer output:
{summary}
* Original task files:
{task_files}

(*@\promptheading{Output Format}@*)
Return ONLY valid JSON with this schema:
{
  "specs": [
    {
      "fine_grained_category":
        "<short kebab-case subtype within {dimension}, e.g. root-and-target-relocation>",
      "concrete_novelty_plan":
        "<the operating-condition change, concretely>",
      "components_to_modify": [
        "environment/Dockerfile",
        "solution/solve.sh",
        "instruction.md",
        "tests/...",
        "harness"
      ],
      "expected_trajectory_change":
        "<how the successful agent run should visibly differ>",
      "why_S_prime_differs_from_S":
        "<why the adapted solution differs from the original>",
      "why_semantics_preserved":
        "<why objective + tests' meaning are unchanged>",
      "task_semantics_preserved": true,
      "minimal_modification_claim": true,
      "implementation_plausibility_score": <1-5>,
      "expected_behavioral_delta_score": <1-5>,
      "risks": [
        "<reasons this novelty may fail>"
      ]
    }
  ]
}

(*@\promptheading{Scoring}@*)
* implementation_plausibility_score: 1 = impossible or too invasive; 3 = requires moderate care; 5 = trivial to implement.
* expected_behavioral_delta_score: 1 = identical solutions; 3 = some visible change; 5 = the original strategy fails.
\end{promptbox}
\FloatBarrier

%% file: sections/appendix/agni-details/prompts/p05-novel-task-instantiator.tex
\begin{promptbox}[prompt:novelty-implementation]{Novel Task Instantiator}
(*@\promptheading{Objective}@*)
Implement the task-preserving novelty for terminal task T. Return the full new content of only the files that change.

(*@\promptheading{Shared Novelty Definition}@*)
{novelty_definition}

(*@\promptheading{Validity Requirements}@*)
1. preserve the same objective;
2. preserve what the tests assert, except for path/invocation updates required by the novelty;
3. be visible in the agent's token stream through commands such as ls, cat, find, which, --version, test errors, permission errors, or runtime behavior;
4. make the unmodified original solve.sh fail or become clearly insufficient under T';
5. include an adapted solve.sh' that succeeds under T';
6. remain realistic and plausible;
7. avoid telling the agent about the novelty in instruction.md unless the original instruction explicitly names something that became invalid.

(*@\promptheading{Inputs}@*)
* Novelty dimension: {dimension}
* Novelty subtype: {subtype}
* Novelty plan:
{plan}
* Original task files:
{task_files}

(*@\promptheading{Requirements}@*)
* Return ONLY valid JSON.
* Emit full new content only for changed files.
* Do not include unchanged files.
* tests/: change only paths/tool invocations/setup assumptions required by the novelty; never weaken or change what is asserted.
* instruction.md: edit only if the original instruction explicitly names a changed path/tool/interface; otherwise leave it unchanged.
* solve.sh: provide the adapted reference solution for T' if solve.sh is part of the task files.
* Keep the privileged novelty truth out of the agent-facing instruction.

(*@\promptheading{Output Format}@*)
Return exactly one JSON object with this schema:
{
  "changed_files": {
    "<relpath>": "<full new file content>"
  },
  "novelty": {
    "dimension": "{dimension}",
    "subtype": "{subtype}",
    "intensity": "mild | moderate | severe",
    "changes": ["<exact concrete change made>"],
    "privileged_spec":
      "<full truth of what changed; not shown to agent>",
    "preservation_argument":
      "<why objective and test meaning are unchanged>",
    "why_original_solution_fails":
      "<why unmodified original solve.sh fails or becomes insufficient>",
    "why_adapted_solution_passes":
      "<why solve.sh' succeeds>",
    "harness_change":
      "none | terminus-2:<what> | echo-minimal:<what>"
  }
}
\end{promptbox}

%% file: sections/appendix/agni-details/prompts/p06-novel-task-instantiation-repair.tex
\begin{promptbox}[prompt:novelty-repair]{Novel Task Instantiation Repair}
(*@\promptheading{Objective}@*)
The novel task T' failed validation: the adapted solve.sh' did not pass the tests, or the build or solution produced an error. Repair the changed files while preserving the intended novelty.

(*@\promptheading{Shared Novelty Definition}@*)
{novelty_definition}

(*@\promptheading{Repair Criteria}@*)
1. the adapted solve.sh' passes in T';
2. the original objective remains unchanged;
3. the tests still assert the same behavior/state;
4. the intended novelty subtype remains present and visible;
5. the unmodified original solve.sh still fails or remains clearly insufficient;
6. changes are minimal and auditable.

(*@\promptheading{Inputs}@*)
* Novelty subtype: {subtype}
* Original task files:
{task_files}
* Current changed files for T':
{current}
* Validation failure output:
{error}

(*@\promptheading{Requirements}@*)
* Return ONLY valid JSON.
* Do not abandon the novelty unless it is impossible; repair it.
* Do not weaken tests to make them pass.
* Change tests only for path/invocation/setup compatibility with the preserved objective.
* If the proposed novelty is impossible to preserve, return "unrepairable": true with a brief reason.

(*@\promptheading{Output Format}@*)
Return exactly one JSON object with this schema:
{
  "changed_files": {
    "<relpath>": "<full new file content>"
  },
  "novelty": {
    "dimension": "<dimension>",
    "subtype": "{subtype}",
    "intensity": "mild | moderate | severe",
    "changes": ["<exact concrete change made>"],
    "privileged_spec":
      "<full truth of what changed; not shown to agent>",
    "preservation_argument":
      "<why objective and test meaning are unchanged>",
    "why_original_solution_fails":
      "<why unmodified original solve.sh fails or becomes insufficient>",
    "why_adapted_solution_passes":
      "<why solve.sh' succeeds>",
    "harness_change":
      "none | terminus-2:<what> | echo-minimal:<what>"
  },
  "unrepairable": false,
  "unrepairable_reason": ""
}
\end{promptbox}

%% file: sections/appendix/agni-details/prompts/p07-adaptive-behavior-validator.tex
\begin{promptbox}[prompt:solution-delta]{Adaptive Behavior Validator}
(*@\promptheading{Objective}@*)
Compare how agents solved the original task T and the novel task T'. Determine whether the novelty changed successful trajectories in the expected way.

(*@\promptheading{Shared Novelty Definition}@*)
{novelty_definition}

(*@\promptheading{Inputs}@*)
* Novelty dimension: {dimension}
* Novelty subtype: {subtype}
* Novelty plan:
{plan}
* Expected trajectory change:
{expected}
* Successful trajectories on T:
{base_traj}
* Successful trajectories on T':
{novelty_traj}

(*@\promptheading{Behavioral Evidence}@*)
Check whether T' caused concrete differences such as:
* different commands
* different files inspected or modified
* discovery before editing
* avoiding removed tools
* handling changed permissions
* ignoring decoys
* adapting to version/runtime errors
* cleanup/idempotence behavior
* changed test/debug loop

(*@\promptheading{Requirements}@*)
* Return ONLY valid JSON.
* Be faithful to trajectories; do not infer differences not visible in logs.
* If trajectories are too sparse to judge, say so.
* A novelty can be valid even if the behavioral delta is weak; distinguish validity from visibility.

(*@\promptheading{Output Format}@*)
Return exactly one JSON object with this schema:
{
  "novelty_visible": true,
  "matches_expected_delta": true,
  "behavioral_delta_strength": "strong | medium | weak | unclear",
  "observed_differences": [
    "<concrete difference seen in T' trajectory>"
  ],
  "unchanged_behaviors": [
    "<important behaviors that stayed the same>"
  ],
  "validity_concerns": [
    "<possible task drift, weak evidence, flaky behavior, etc.>"
  ],
  "rationale": "<1-3 sentences>"
}
\end{promptbox}

%% file: sections/appendix/agni-details/prompts/p08-trajectory-adaptation-analyzer.tex
\FloatBarrier
\begin{promptbox}[prompt:trajectory-analysis]{Trajectory Adaptation Analyzer}
(*@\promptheading{Role}@*)
You are an expert in terminal tasks, shell behavior, command-line debugging, and agent-trajectory analysis.

(*@\promptheading{Objective}@*)
Analyze each step of an agent trajectory in a novel environment. Summarize the step's main semantic intent, assign every substantive intent class, judge the outcome of each intent from visible evidence, and label progress in handling the environmental novelty. Do not summarize commands individually or invent unsupported details.

(*@\promptheading{Novelty-Handling Stages}@*)
Assign zero or more stages to each step:
* observed: the observation newly exposes direct evidence of the novelty or a clearly novelty-attributable consequence.
* recognized: the agent explicitly treats prior evidence as unexpected, suspicious, or inconsistent with the task.
* diagnosed: the agent identifies the effective changed condition at the level needed to adapt.
* revised: the step successfully removes or bypasses the novelty obstacle, but the complete task remains unfinished or incorrect.
* solved: the step completes or verifies the full task under the novelty using a task-preserving method.

Use [] when no stage applies. Do not assign observed from hidden provenance or grader information, and do not mark the same evidence observed again unless the step reveals something new. Evidence exposed in step N's observation may be labeled recognized or diagnosed only when the agent subsequently engages with it. A failed attempted revision is not revised. Assign solved only when the complete task state is produced or verified by a task-preserving method. Deterministic command semantics may support solved even when the intent outcome is inconclusive because exact post-hoc verification is absent. An exploit may still be observed, recognized, or diagnosed, but it is never solved.

(*@\promptheading{Intent Classes}@*)
Assign all substantive intents present in each agent step:
* inspect: examine ordinary task inputs, files, or requirements before attempting a solution.
* execute: make the first substantive task-producing attempt.
* diagnose: investigate the cause of a visible failure or anomaly.
* explore: look for another tool, file, path, resource, or method without applying it yet.
* retry: repeat the same failed method or a superficial variant that preserves the failed assumption.
* revise: apply a task-preserving changed method intended to overcome the novelty.
* repair: finish, format, permission-fix, regenerate, or clean up required artifacts after the novelty obstacle is handled.
* verify: check existing outputs against task requirements without materially changing them.
* finalize: request or confirm completion without making substantive environment changes.
* exploit: satisfy visible checks while replacing or violating the required task semantics.
* other: use only when no listed class applies, and explain why.

Write one intent_summary per step and one intent-analysis object per assigned class. When reasoning and actions disagree, classify the actions and note the mismatch. Ignore incidental setup, logging, or checks that do not affect the step's main purpose.

Use execute for the first substantive attempt even when it is already robust to the novelty. After a relevant failure, use retry when the new attempt preserves the failed assumption and revise only when it changes the novelty-relevant method. An unrelated typo, quoting error, or setup mistake does not by itself turn the next real attempt into retry. A step may receive multiple labels, such as diagnose and revise, when both behaviors are substantive. After the novelty obstacle is handled, use repair for remaining artifact work rather than revise.

(*@\promptheading{Intent Outcomes}@*)
For each assigned intent, choose success, partial, failure, or inconclusive using the step's observation and visible deterministic command semantics. Judge each intent separately.

For verify, instead use:
* verify_valid: the check covers the relevant criteria and the artifact is correct.
* verify_invalid: the check correctly establishes that the artifact is wrong.
* verify_partial: the check covers only part of the relevant criteria.
* verify_inconclusive: the check fails, tests the wrong requirement, or provides no usable verdict.

For diagnose, success means identifying the effective cause and partial means narrowing it to a useful symptom or component. For explore, success means finding a usable alternative. An expected failing subcommand inside an explicit fallback does not by itself make the intent fail; judge whether the fallback succeeds.

For finalize only, the grader result may be used together with the observation. A grader pass obtained through an exploit is still a successful finalization, but it is not a genuine novelty fix.

(*@\promptheading{Grounding Rules}@*)
* Base every per-step label and outcome on that step's reasoning, actions, observation, or deterministic command semantics.
* Use the novelty description, original solution, and grader result only as context; do not backfill per-step progress from them.
* A generic error, missing file, or empty output is not observed unless the evidence specifically connects it to the changed condition.
* Do not infer recognized or diagnosed merely because terminal output contains a clue; the agent must engage with it.
* Assign revise only when the changed method addresses the novelty-relevant assumption, not for superficial flag, path, ordering, or quoting changes.
* Assign exploit only when the action replaces the genuine required operation to satisfy a check. A different method that genuinely achieves the required state is not an exploit.
* An exploit is revised only if it genuinely clears or bypasses the novelty obstacle; merely fabricating evaluator-visible artifacts does not count as revision.
* Do not treat completion claims, completion-tool calls, or superficial checks as proof that the task is solved.
* When exactness matters and the relevant output is truncated or unverified, use inconclusive.
* Do not invent task requirements, constraints, tests, or failure causes absent from the visible inputs.

(*@\promptheading{Constraint Violations}@*)
Separately list stated task constraints that the agent visibly violates, such as privilege, allowed-directory, forbidden-operation, package-source, or required-path constraints. Do not infer unstated constraints. Ordinary commands executed from a root shell do not alone violate a no-root constraint. Temporary use of /tmp or /var/tmp does not violate an allowed-directory constraint unless a required artifact is placed outside the allowed location. Set grader_enforced to false if the grader passes despite the violation. Constraint violations do not change intent outcomes or novelty-handling stages. Use [] when no stated constraint is violated.

(*@\promptheading{Output Format}@*)
Return ONLY valid JSON with this structure:
{
  "task_goal": "<one or two sentences>",
  "constraint_violations": [
    {
      "step_id": 3,
      "constraint": "<stated constraint>",
      "violation": "<visible violation>",
      "grader_enforced": false
    }
  ],
  "step_analysis": [
    {
      "step_id": 1,
      "actor": "user",
      "intent_summary": "<summary of the task instruction>",
      "intent_analysis": [],
      "intent_outcomes": [],
      "novelty_handling_analysis": [],
      "novelty_handling_comment": "<why no novelty label applies>",
      "comment": ""
    },
    {
      "step_id": 2,
      "actor": "agent",
      "intent_summary": "<main subgoal and semantic strategy>",
      "intent_analysis": [
        {
          "intent_reasoning": "<evidence for this intent>",
          "intent_class": "execute"
        }
      ],
      "intent_outcomes": [
        {
          "intent_class": "execute",
          "outcome_justification": "<visible evidence for the outcome>",
          "outcome": "success"
        }
      ],
      "novelty_handling_analysis": [
        {
          "reason": "<visible evidence for the stage>",
          "label": "observed"
        }
      ],
      "novelty_handling_comment": "<why no additional stages apply>",
      "comment": "<optional ambiguity or reasoning-action mismatch>"
    }
  ],
  "trajectory_summary": "<concise account of novelty handling over time>"
}

(*@\promptheading{Output Requirements}@*)
* step_id must match the trajectory, and actor must be user or agent.
* User-only instruction steps have empty intent_analysis and intent_outcomes arrays.
* Each agent intent has one intent_analysis object and one matching intent_outcomes object.
* Write intent_reasoning before intent_class, and outcome_justification before outcome.
* Every intent_class is one of inspect, execute, diagnose, explore, retry, revise, repair, verify, finalize, exploit, or other.
* novelty_handling_analysis contains zero or more objects with reason before label.
* Every novelty label is one of observed, recognized, diagnosed, revised, or solved.
* Ground every justification in visible evidence or deterministic command semantics.
* Return no Markdown fences, comments, or prose outside the JSON.

(*@\promptheading{Inputs}@*)
Task instruction:
{{original_task}}

Novel environment details:
{{shift_description}}

Original successful solution S:
{{original_solution}}

Agent trajectory in the novel environment:
{{agent_trajectory}}

Grader result (context only, except for finalization outcome):
{{grader_result}}
\end{promptbox}
\FloatBarrier

%% file: sections/appendix/agni-details/prompts/p09-novelty-disclosure-generator.tex
\begin{promptbox}[prompt:disclose-novelty]{Novelty Disclosure Generator}
(*@\promptheading{Role}@*)
You are preparing a novelty-disclosed version of a terminal-task instruction for an agentic evaluation.

(*@\promptheading{Context}@*)
The task environment has been modified in a specific, non-standard way. The task objective and required final state are unchanged. In the standard condition, the agent is not told about the modification and must discover it. In this disclosed condition, the agent is told exactly what is non-standard, but not how to work around, fix, or adapt to it.

(*@\promptheading{Objective}@*)
Write a short environment note of 2-6 sentences in plain prose to append to the task instruction.

(*@\promptheading{Requirements}@*)
* State clearly and factually what is non-standard: identify the modified component and explain how it behaves differently from the standard environment.
* Describe only the environmental change and its observable behavior. Do not restate the task objective or required final state.
* Never reveal, recommend, or hint at a workaround, fix, alternative tool, command, flag, path, or adaptation strategy. Do not say "use X instead," "you should," or "to handle this." Ignore any adaptation or solution guidance in the internal description.
* Do not mention the reference solution, tests, graders, benchmarks, scoring, or that a novelty was applied. Write a neutral factual notice about the environment.

(*@\promptheading{Inputs}@*)
* Original task instruction:
{instruction}
* Environmental modification (disclose the change, never the fix):
{novelty_changes}

(*@\promptheading{Output Format}@*)
Output ONLY the environment-note prose. Do not include a heading, preamble, or Markdown fence.
\end{promptbox}

%% file: sections/appendix/agni-details/prompts/p10-novelty-realism-judge.tex
\FloatBarrier
\begin{promptbox}[prompt:realism-judgment]{Novelty Realism Judge}
(*@\promptheading{Role}@*)
You are auditing the realism of a synthetic environmental novelty injected into a terminal-based software-agent task. The original task T was changed into T' by modifying only the environment. The goal remains unchanged, but the agent's previously successful strategy should fail and require adaptation.

(*@\promptheading{Objective}@*)
Evaluate two levels independently:
* Underlying condition: the semantic environmental change that invalidates the prior strategy, such as a tool interface differing from the assumed version.
* Realization mechanism: how that condition is concretely instantiated, such as a different installed version, modified CLI, or incompatible executable earlier in PATH.

An observable symptom is evidence about the realization, not a separate dimension. Follow this order: identify the underlying condition, score its occurrence plausibility, score realization fidelity, and classify the implementation mechanism.

(*@\promptheading{A. Underlying Condition}@*)
Infer the underlying condition from the supplied artifacts and state it independently of the benchmark-specific implementation. Cite the files, paths, patch hunks, or interfaces that establish it. Do not alter the inferred condition to justify a preferred score.

Judge from the diff and changed files rather than the generator's claims. When the artifacts contradict or narrow the claimed condition, the artifacts win. For example, if the specification claims a general version change but the diff contains a wrapper that special-cases one argument sequence, evaluate the narrower implemented behavior.

(*@\promptheading{B. Occurrence Plausibility}@*)
Question: could the underlying condition reasonably arise during real deployment or maintenance of a terminal-based software agent?

1. Implausible: no credible real-world mechanism; mainly an evaluator trick or behavior real systems would not reasonably exhibit.
2. Technically possible but contrived: requires an artificial setup, unlikely coincidence, or mechanism unrepresentative of ordinary deployment.
3. Plausible in some deployments: has a credible mechanism but is uncommon, domain-specific, or somewhat stylized.
4. Clearly realistic: a recognizable deployment, configuration, dependency, permission, filesystem, interface, or runtime change encountered by practitioners.
5. Directly representative: closely matches a documented, observed, or widely recognized deployment failure mode strongly supported by the artifacts.

Do not assign 5 merely because the condition is imaginable. Score the condition itself, independent of implementation quality, reference solution, or recovery strategy.

(*@\promptheading{C. Realization Fidelity}@*)
Question: does the concrete implementation faithfully reproduce how the condition would manifest in a real system?

1. Unfaithful or incoherent: contradicts the condition, relies mainly on evaluator-only machinery, or changes the task through an unrelated mechanism.
2. Crude proxy: captures the broad idea but introduces major artificial artifacts or substantially different behavior.
3. Controlled abstraction: preserves the central causal effect and relevant adaptation but simplifies meaningful aspects coherently.
4. Faithful simulation: uses realistic mechanisms and produces deployment-like observations and consequences, with only minor simplification or scaffolding.
5. Direct or near-exact reproduction: directly instantiates the real mechanism, or reproduces its behavior, scope, consequences, and recovery options with negligible abstraction.

Judge the realization from the diff and changed files. A wrapper, shim, or fixture is not automatically unfaithful; judge whether it preserves the causal mechanism and agent-visible behavior. A wrapper that special-cases one exact command or anticipated solution is generally a crude or controlled proxy, not a faithful simulation. Do not raise the score merely because the goal is preserved, the task is solvable, or recovery exists.

(*@\promptheading{Required Evidence}@*)
For the condition and both scores, cite concrete evidence such as a file, path, patch hunk, permission, command interface, package or runtime version, error behavior, or configuration value. Explain:
* what condition the shift instantiates;
* how that condition could arise in real systems; and
* whether the concrete implementation matches that mechanism.

Distinguish observed evidence from assumptions. Do not use generic claims such as "such issues happen in practice." The reference solution, tests, and grader are withheld; judge only from the environment construction and agent-facing instruction.

(*@\promptheading{D. Mechanism Classification}@*)
Classify the implementation mechanism independently of the two scores:
* Choose exactly one mechanism family copied verbatim from the supplied taxonomy, or other if none fits.
* For other, provide a plain-language family name and representative example in proposed_family_if_other.
* Classify the realization visible in the diff, not the specification's claimed dimension.
* adaptive_cluster must exactly match a taxonomy cluster. For other, choose the cluster whose recovery loop best matches what the agent must do.
* surface_binding must name the concrete surface touched, such as a tool, binary, path, environment variable, file format, or encoding.

Taxonomy:
{taxonomy}

(*@\promptheading{Inputs}@*)
* Task ID: {task_id}
* Agent-facing instruction for T':
{instruction}
* Generator's novelty specification:
{spec}
* Unified diff from T to T':
{diff}
* Full contents of changed files in T':
{changed_files}

(*@\promptheading{Output Format}@*)
Return ONLY one JSON object with exactly this shape and no extra keys or Markdown. Within each scored dimension, place evidence and rationale before the score.

Allowed confidence values are low, medium, and high.

{
  "task_id": "{task_id}",
  "change_summary": "<one or two sentences describing the environmental change>",
  "underlying_condition": "<semantic condition stated without implementation details>",
  "occurrence_plausibility": {
    "condition_identification_evidence": [
      {
        "source": "<file, path, hunk, or interface>",
        "observation": "<what the artifact shows>",
        "implication": "<what condition this establishes>"
      }
    ],
    "real_world_basis": "<how the condition could arise in real systems>",
    "rationale": "<why the mechanism is or is not credible in deployment>",
    "score": 1,
    "confidence": "low"
  },
  "realization_fidelity": {
    "evidence": [
      {
        "source": "<file, path, hunk, or interface>",
        "observation": "<what the artifact shows>",
        "implication": "<what this establishes about the realization>"
      }
    ],
    "rationale": "<how faithfully the realization implements the condition>",
    "score": 1,
    "confidence": "low"
  },
  "mechanism": {
    "rationale": "<why the selected family matches, or why none matches>",
    "mechanism_family": "<verbatim taxonomy family, or other>",
    "proposed_family_if_other": "<name and example, or empty if not other>",
    "adaptive_cluster": "<verbatim taxonomy cluster>",
    "surface_binding": "<specific tool, path, variable, format, or encoding>",
    "confidence": "low"
  }
}
\end{promptbox}
\FloatBarrier